\documentclass{article}
\usepackage{natbib}

\usepackage{xcolor} 
\usepackage{graphicx}
\usepackage{hyperref}
\usepackage{listings}
\usepackage{amsmath}
\usepackage{amsfonts}
\usepackage{MnSymbol}

\newtheorem{THEOREM}{Theorem}[section]
\newenvironment{theorem}{\begin{THEOREM} \hspace{-.85em} {\bf :} }%
                        {\end{THEOREM}}
\newtheorem{LEMMA}[THEOREM]{Lemma}
\newenvironment{lemma}{\begin{LEMMA} \hspace{-.85em} {\bf :} }%
                      {\end{LEMMA}}
\newtheorem{COROLLARY}[THEOREM]{Corollary}
\newenvironment{corollary}{\begin{COROLLARY} \hspace{-.85em} {\bf :} }%
                          {\end{COROLLARY}}
\newtheorem{PROPOSITION}[THEOREM]{Proposition}
\newenvironment{proposition}{\begin{PROPOSITION} \hspace{-.85em} {\bf :} }%
                            {\end{PROPOSITION}}
\newtheorem{DEFINITION}[THEOREM]{Definition}
\newenvironment{definition}{\begin{DEFINITION} \hspace{-.85em} {\bf :} \rm}%
                            {\end{DEFINITION}}
\newtheorem{CLAIM}[THEOREM]{Claim}
\newenvironment{claim}{\begin{CLAIM} \hspace{-.85em} {\bf :} \rm}%
                            {\end{CLAIM}}
\newtheorem{EXAMPLE}[THEOREM]{Example}
\newenvironment{example}{\begin{EXAMPLE} \hspace{-.85em} {\bf :} \rm}%
                            {\end{EXAMPLE}}
\newtheorem{REMARK}[THEOREM]{Remark}
\newenvironment{remark}{\begin{REMARK} \hspace{-.85em} {\bf :} \rm}%
                            {\end{REMARK}}

\newcommand{\thm}{\begin{theorem}}
\newcommand{\lem}{\begin{lemma}}
\newcommand{\pro}{\begin{proposition}}
\newcommand{\dfn}{\begin{definition}}
\newcommand{\rem}{\begin{remark}}
\newcommand{\xam}{\begin{example}}
\newcommand{\cor}{\begin{corollary}}
\newcommand{\prf}{\noindent{\bf Proof:} }
\newcommand{\ethm}{\end{theorem}}
\newcommand{\elem}{\end{lemma}}
\newcommand{\epro}{\end{proposition}}
\newcommand{\edfn}{\bbox\end{definition}}
\newcommand{\erem}{\bbox\end{remark}}
\newcommand{\exam}{\bbox\end{example}}
\newcommand{\ecor}{\end{corollary}}
\newcommand{\eprf}{\bbox\vspace{0.1in}}
\newcommand{\beqn}{\begin{equation}}
\newcommand{\eeqn}{\end{equation}}

\newcommand{\bbox}{\vrule height7pt width4pt depth1pt}

\newcommand{\clm}{\begin{claim}}
\newcommand{\eclm}{\end{claim}}

\newcommand{\sat}{\models}

\newcommand{\union}{\cup}

\renewcommand{\phi}{\varphi}

\newcommand{\F}{{\cal F}}

\newcommand{\R}{{\cal R}}

\newcommand{\U}{{\cal U}}
\newcommand{\V}{{\cal V}}

\newcommand{\ol}{\setlength{\itemsep}{0pt}\begin{enumerate}}
\newcommand{\eol}{\end{enumerate}\setlength{\itemsep}{-\parsep}}
\newcommand{\ul}{\setlength{\itemsep}{0pt}\begin{itemize}}
\newcommand{\dl}{\setlength{\itemsep}{0pt}\begin{description}}
\newcommand{\edl}{\end{description}\setlength{\itemsep}{-\parsep}}
\newcommand{\eul}{\end{itemize}\setlength{\itemsep}{-\parsep}}

\newcommand{\commentout}[1]{}

\newcommand{\bi}{\begin{itemize}}
\newcommand{\ei}{\end{itemize}}
\newcommand{\be}{\begin{enumerate}}
\newcommand{\ee}{\end{enumerate}}

\newcommand{\noproofs}{\commentout}
\newcommand{\proofs}{}

\newtheorem{principle}{Principle}

\newtheorem{observation}{Observation}

\newenvironment{oldthm}[1]{\par\noindent{\bf Theorem #1:} \em \noindent}{\par}
\newenvironment{oldlem}[1]{\par\noindent{\bf Lemma #1:} \em \noindent}{\par}
\newenvironment{oldcor}[1]{\par\noindent{\bf Corollary #1:} \em \noindent}{\par}
\newenvironment{oldpro}[1]{\par\noindent{\bf Proposition #1:} \em \noindent}{\par}
\newcommand{\othm}[1]{\begin{oldthm}{\ref{#1}}}
\newcommand{\eothm}{\end{oldthm} \medskip}
\newcommand{\olem}[1]{\begin{oldlem}{\ref{#1}}}
\newcommand{\eolem}{\end{oldlem} \medskip}
\newcommand{\ocor}[1]{\begin{oldcor}{\ref{#1}}}
\newcommand{\eocor}{\end{oldcor} \medskip}
\newcommand{\opro}[1]{\begin{oldpro}{\ref{#1}}}
\newcommand{\eopro}{\end{oldpro} \medskip}

\begin{document}

\title{Causal Sufficiency and Actual Causation \\ \footnotesize{(Preprint of paper to appear in the {\em Journal of Philosophical Logic})}}
\author{Sander Beckers \\ 
Munich Center for Mathematical Philosophy, LMU Munich \\ 
srekcebrednas@gmail.com}
\date{}

\maketitle

\begin{abstract} 
Pearl opened the door to formally defining {\em actual causation} using causal models. His approach rests on two strategies: first, capturing the widespread intuition that  $X=x$ causes $Y=y$ iff $X=x$ is a {\em Necessary Element of a Sufficient Set} for $Y=y$, and second, showing that his definition gives intuitive answers on a wide set of problem cases. This inspired dozens of variations of his definition of actual causation, the most prominent of which are due to Halpern \& Pearl. Yet all of them ignore Pearl's first strategy, and the second strategy taken by itself is unable to deliver a consensus. This paper offers a way out by going back to the first strategy: it offers six formal definitions of {\em causal sufficiency} and two interpretations of necessity. Combining the two gives twelve new definitions of actual causation. Several interesting results about these definitions and their relation to the various Halpern \& Pearl definitions are presented. Afterwards the second strategy is evaluated as well. In order to maximize neutrality, the paper relies mostly on the examples and intuitions of Halpern \& Pearl. One definition comes out as being superior to all others, and is therefore suggested as a new definition of actual causation.
\end{abstract}

\section{Introduction}

Two decades have passed since Judea Pearl's groundbreaking book on causality was published \citep{pearl:book}. It offers a formal account of causal models that led causal modeling to become a central part of Artificial Intelligence. One of the book's most important applications for philosophy is its formal definition of {\em actual causation}, i.e., causation of particular events. 

Pearl defends his account of actual causation using two strategies. The first strategy starts with the widely shared intuition that $X=x$ causes $Y=y$ iff $X=x$ is a {\em Necessary Element of a Sufficient Set} for $Y=y$ (the NESS intuition, from now on).\footnote{This acronym was coined by \cite{wright88}, but Pearl does not intend to formalize the specific manner in which Wright understood it, nor do I in the current paper. I have formalized Wright's interpretation of the NESS definition elsewhere, in the process of developing another definition of causation \citep{cness}. The latter definition is in many ways a simplification of the definition that I defend here. The precise relation between these two definitions is the subject of future work.}\footnote{\cite{mackie65} formulates the same intuition differently, resulting in the equally famous INUS acronym. See \cite{wright11} for a detailed discussion of the subtle differences between them.} Pearl claims that using causal models allows one to make this intuition formally precise, whereas existing logical notions of necessity and sufficiency lack the resources to do so. The second strategy is to demonstrate that his formal account offers intuitive verdicts for a number of problematic examples. 

Ever since, Pearl's account has come under severe criticism. By now there are dozens of papers -- both from philosophers and from researchers in AI -- attempting to improve upon his account.\footnote{Just to name some of the most influential ones: \cite{hitchcock01,hitchcock07,woodward,hall07,weslake}.} Most prominently, Pearl himself has offered several revisions of his account in collaboration with Halpern, culminating in the most recent revision by Halpern individually \citep{pearl:book2,halpernpearl01,halpernpearl05a,halpern15,halpernbook}. Together these accounts of causation are referred to as the Halpern \& Pearl definitions, or {\em HP definitions} for short, and they are by far the most influential accounts of causation out there.

The problem with all of these attempts at revising Pearl's initial account, is that they completely ignore the first strategy and focus almost excusively on the second strategy. Roughly put, the typical setup is to go over some examples for which existing definitions give counterintuitive answers, and then to construct a new definition that does not do so. It is unrealistic to expect that this second strategy in and of itself can deliver a satisfactory account of causation, because there are too many  examples and even more intuitions \citep{stonesoup,beckers18a}. 

To solve this problem, this paper starts out with an explicit focus on the first strategy. It is striking that immediately after discussing the NESS intuition, Pearl diverges into complicated technical notions like ``sustenance'' and ``causal beams'' and never looks back, be it in his book or in the subsequent work on the HP definitions. Instead I offer what is the most natural route down the first strategy, namely to look at formalizations of {\em causal sufficiency} (as opposed to logical sufficiency) and combine them with two interpretations of {\em necessity}. Taken together this results in twelve distinct formal definitions of actual causation. 

These definitions are compared to each other and to the HP definitions, leading to several interesting results. For one, it turns out that one of these twelve definitions is equivalent to the most recent HP definition \citep{halpern15,halpernbook}. Therefore this paper is the first to show that one of the HP definitions succeeds in delivering Pearl's promise. At the same time, it also shows that the other HP definitions do not. 

Next we turn to the second strategy. Given the diversity of intuitions about the many examples presented in the literature, the best we can do is arrive at a comparative verdict: does one of the definitions here developed fare better than the HP definitions? In order to avoid relying on my own intuitions, I present two criteria by which we can answer this question. First, I make use of Halpern and Pearl's own examples and rely almost exclusively on their intuitions, which for the most part align with the consensus in the literature. (Example \ref{ex:noname} forms a notable exception.) Here the answer is that one of the twelve definitions does better than the HP definitions. Second, I present six examples that are very similar to each other, and assess which definitions are able to handle them in a consistent (and preferably also intuitive) manner. Here the answer is that the previous definition again does better than the HP definitions.

Therefore I suggest adopting this definition of actual causation. Roughly, this definition states that $X=x$ causes $Y=y$ iff there is a set $\vec{W}=\vec{w}$ so that $(X=x,\vec{W}=\vec{w})$ is sufficient for $Y=y$ along a causal network $\vec{N}$ {\em and} there exists some value $x'$ so that $(X=x',\vec{W}=\vec{w})$ is not sufficient for $Y=y$ along any causal subnetwork of $\vec{N}$. 

This paper is laid out as follows. The next section introduces {\em structural equations models}, the formal causal models that are used to express all the definitions. Then I state the three most recent HP definitions in Section \ref{sec:hp}. Section \ref{sec:suf} presents six notions of causal sufficiency and shows how they relate to each other. We then use these six notions to formalize actual causation along the NESS intuition in Section \ref{sec:ac}, and discuss several interesting results. After this theoretical groundwork, we start looking for the best definition. Two definitions are discarded by showing that they have certain unacceptable properties in Section \ref{sec:4and6}. Finally, Section \ref{sec:237} compares the remaining definitions to each other and to the HP definitions by considering examples from Halpern \& Pearl and a few additional ones.

\section{Structural Equations Modeling}\label{sec:equations}

This section reviews the definition of causal models as they were introduced by \cite{pearl:book}.
Much of the discussion and notation is taken from \cite{halpernbook} with little change.

\begin{definition}
A signature $\cal S$ is a tuple $(\U,\V,\R)$, where $\U$
is a set of \emph{exogenous} variables, $\V$ is a set 
of \emph{endogenous} variables,
and $\R$ a function that associates with every variable $Y \in  
\U \union \V$ a nonempty set $\R(Y)$ of possible values for $Y$
(i.e., the set of values over which $Y$ {\em ranges}).
If $\vec{X} = (X_1, \ldots, X_n)$, $\R(\vec{X})$ denotes the
crossproduct $\R(X_1) \times \cdots \times \R(X_n)$.
\end{definition}

Exogenous variables represent factors whose causal origins are outside the scope of the causal model, such as background conditions and noise. 
The values of the endogenous variables, on the other hand, are causally determined by other variables within the model (both endogenous and exogenous). 

\begin{definition}
A \emph{causal model} $M$ is a pair $(\cal S,\F)$, 
where $\cal S$ is a signature and
$\F$ defines a  function that associates with each endogenous
variable $X$ a \emph{structural equation} $F_X$ giving the value of
$X$ in terms of the 
values of other endogenous and exogenous variables. Formally, the equation $F_X$ maps $\R(\U \union \V - \{X\})$  to $\R(X)$,
so $F_X$ determines the value of $X$, 
given the values of all the other variables in $\U \union \V$.  
\end{definition}

Note that there are no functions associated with exogenous variables;
their values are determined outside the model.  We call a setting
$\vec{u} \in \R(\U)$ of values of exogenous variables a \emph{context}.

The value of $X$ may depend on the values of only a few other variables.  
$X$ \emph{depends on $Y$ in context $\vec{u}$}
if there is some setting of the endogenous variables  
other than $X$ and $Y$ such that 
if the exogenous variables have value $\vec{u}$, then varying the value
of $Y$ in that 
context results in a variation in the value of $X$; that 
is, there is a setting $\vec{z}$ of the endogenous variables other than $X$ and
$Y$ and values $y$ and $y'$ of $Y$ such that $F_X(y,\vec{z},\vec{u}) \ne
F_X(y',\vec{z},\vec{u})$. We then say that $Y$ is a {\em parent} of $X$. 

We extend this genealogical terminology in the usual manner, by taking the {\em ancestor} relation 
to be the transitive closure of the parent relation (i.e., $Y$ is an ancestor of $X$ iff there exist variables so that $Y$ is a parent of $V_1$, $V_1$ is a parent of $V_2$, ..., and $V_n$ is a parent of $X$). The {\em descendant} relation is simply the reversal of the ancestor relation (i.e., $X$ is a descendant of $Y$ iff $Y$ is an ancestor of $X$.) A {\em path} is a sequence of variables in which each element is a child of the previous element.

In this paper we restrict attention to \emph{strongly recursive} (or
\emph{strongly acyclic}) models, that is, models where there is a partial order
$\preceq$ on variables such that if $Y$ depends on $X$, then $X \prec Y$. 
In a strongly recursive model, given a context $\vec{u}$, 
the values of all the remaining variables are determined (we can just
solve for the value of the variables in the order given
by $\preceq$). We often write the
equation for an endogenous variable as $X = f(\vec{Y})$; this denotes
that the value of $X$ depends only on the values of the variables in
$\vec{Y}$, and the connection is given by the function $f$. For example, we might
have $X = Y + 5$. 
 
An \emph{intervention} has the form $\vec{X} \gets \vec{x}$, where $\vec{X}$ 
is a set of endogenous variables.  Intuitively, this means that the
values of the variables in $\vec{X}$ are set to the values $\vec{x}$.  
The structural equations define what happens in the presence of 
interventions.  Setting the value of some variables $\vec{X}$ to
$\vec{x}$ in a causal 
model $M = (\cal S,\F)$ results in a new causal model, denoted $M_{\vec{X}
\gets \vec{x}}$, which is identical to $M$, except that $\F$ is
replaced by $\F^{\vec{X} \gets \vec{x}}$: for each variable $Y \notin
  \vec{X}$, $F^{\vec{X} \gets \vec{x}}_Y = F_Y$ (i.e., the equation
  for $Y$ is unchanged), while for
each $X'$ in $\vec{X}$, the equation $F_{X'}$ for $X'$ is replaced by $X' = x'$
(where $x'$ is the value in $\vec{x}$ corresponding to $X'$).

Given a signature $\cal S = (\U,\V,\R)$, an \emph{atomic formula} is a
formula of the form $X = x$, for  $X \in \V$ and $x \in \R(X)$.  
A {\em causal formula (over $\cal S$)\/} is one of the form
$[Y_1 \gets y_1, \ldots, Y_k \gets y_k] \phi$, where
\begin{itemize}
\item $\phi$ is a Boolean combination of atomic formulas,
\item $Y_1, \ldots, Y_k$ are distinct variables in $\V$, and
\item $y_i \in \R(Y_i)$ for each $1 \leq i \leq k$.
\end{itemize}
Such a formula is abbreviated as $[\vec{Y} \gets \vec{y}]\phi$. The special case where $k=0$ is abbreviated as $\phi$.
Intuitively, $[Y_1 \gets y_1, \ldots, Y_k \gets y_k] \phi$ says that $\phi$ would hold if $Y_i$ were set to $y_i$, for $i = 1,\ldots,k$.

A causal formula $\psi$ is true or false in a \emph{causal setting}, which is a causal model given a
context. As usual, we write $(M,\vec{u}) \sat \psi$  if the causal
formula $\psi$ is true in the causal setting $(M,\vec{u})$.
The $\sat$ relation is defined inductively. 
$(M,\vec{u}) \sat X = x$ if the variable $X$ has value $x$
in the unique (since we are dealing with recursive models) solution
to the equations in $M$ in context $\vec{u}$ (i.e., the unique vector
of values that simultaneously satisfies all
equations in $M$ with the variables in $\U$ set to $\vec{u}$).
The truth of conjunctions and negations is defined in the standard way.
Finally, $(M,\vec{u}) \sat [\vec{Y} \gets \vec{y}]\phi$ if 
$(M_{\vec{Y} \gets \vec{y}},\vec{u}) \sat \phi$ (i.e., the intervention $\vec{Y} \gets \vec{y}$ transforms $M$ into a new model $M_{\vec{Y} \gets \vec{y}}$, in which we assess the truth of $\phi$).

\section{HP Definitions}\label{sec:hp}

Now on to the HP definitions. As \cite{pearl:book}'s initial definition is a precursor to the HP definitions that gives less intuitive results and is far more complicated, I do not discuss it. (It is safe to say that by now it has been unanimously rejected.) Two of the HP definitions are developed by both Halpern and Pearl, whereas the third one is solely due to Halpern. The relations between them are extensively discussed by \cite{halpernbook}. 

The general form of all three definitions is as follows (where $\phi$ is a Boolean combination of atomic formulas):
\begin{definition}\label{def:HP} 
$\vec{X} = \vec{x}$ is an \emph{actual cause} of $\phi$ in  $(M,\vec{u})$ if the following three conditions hold: 
\begin{description}
\item[{\rm AC1.}]\label{ac1} $(M,\vec{u}) \sat (\vec{X} =
  \vec{x}) \land \phi$. 
\item[{\rm AC2.}] See below. 
\item[{\rm AC3.}] \label{ac3}\index{AC3}  
  $\vec{X}$ is minimal; there is no strict subset $\vec{X}''$ of
    $\vec{X}$ such that $\vec{X}'' = \vec{x}''$ satisfies
AC2, where $\vec{x}''$ is the restriction of
$\vec{x}$ to the variables in $\vec{X}''$.
\end{description}
\end{definition}

Questions of actual causation are posed relative to an {\em actual context} $\vec{u}$, because as we know from the previous section a context completely determines which events actually took place. So AC1 represents the trivial requirement that the candidate cause and effect are among the events which took place. AC3 is also fairly straightforward: we should not consider redundant elements to be parts of causes. The real content of the definition lies with AC2. 

Throughout the rest of the paper, settings of variables $\vec{V}$ with superscript $^*$ (i.e., $\vec{v}^*$) indicate that $(M,\vec{u}) \sat (\vec{V} =  \vec{v}^*)$. Settings of variables $\vec{V}$ with superscript $'$ (i.e., $\vec{v}'$) indicate that $(M,\vec{u}) \sat (V \neq  v')$ for each $V \in \vec{V}$. Settings of variables without any superscript can refer to any setting.

In line with the NESS intuition, we should expect AC2 to consist of formal variants of these two conditions:\footnote{I list them unalphabetically for consistency with the HP definitions.}
\begin{description}
\item[{\rm AC2(b).}] There is a set $\vec{W}$ so that $(\vec{X}=\vec{x},\vec{W}=\vec{w}^*)$ is causally sufficient for $\phi$.
\item[{\rm AC2(a).}] $\vec{X}=\vec{x}$ is necessary for the sufficiency of $(\vec{X}=\vec{x},\vec{W}=\vec{w}^*)$.
\end{description}

At first glance, the first two HP definitions seem to meet this expectation: they consist of conditions AC2(a) and AC2(b), and Halpern refers to these as a ``necessity condition'' and a ``sufficiency condition'' (\citeyear[p. 3]{halpern15}). Upon closer examination, however, it is hard to see how either version of AC2(b) can sensibly be interpreted as capturing causal sufficiency. 

We start with {\bf Original HP} \citep{halpernpearl01}:
\begin{definition}\label{def:OHP} [{\bf Original HP}]
\begin{description}
\item[{\rm AC2(a).}] There is a partition of $\V$ into two sets $\vec{Z}$ and $\vec{W}$ with $\vec{X} \subseteq \vec{Z}$ and a setting $\vec{x}'$ and $\vec{w}$ of the variables in $\vec{X}$ and $\vec{W}$, respectively, such that 
$(M,\vec{u}) \sat  [\vec{X} \gets \vec{x}', \vec{W} \gets \vec{w}] \neg \phi.$
\item[{\rm AC2(b).}] For all subsets $\vec{Y}$ of $\vec{Z} \minus \vec{X}$, we have $(M,\vec{u}) \sat  [\vec{X} \gets \vec{x}, \vec{W} \gets \vec{w}, \vec{Y} \gets \vec{y}^*] \phi.$
\end{description}
We call $\vec{W}=\vec{w}$ a {\em witness} of $\vec{X}=\vec{x}$ causing $Y=y$.
\end{definition}

\commentout{
If we ignore $\vec{Z}$ for a moment, then AC2 can be read as stating that the effect counterfactually depends on the cause holding fixed a certain witness $\vec{W}=\vec{w}$. Under this reading, AC2(a) can be interpreted as offering a contrastive notion of necessity: there exist contrast values for $\vec{X}$ so that the sufficiency condition expressed in AC2(b) no longer holds. Yet there is no obvious way in which this simplified view allows AC2(b) to be interpreted as capturing causal sufficiency, let alone a way in which the full condition that includes $\vec{Z}$ and its subsets can be so interpreted. This is confirmed by the fact that Halpern \& Pearl do not even offer an attempt at giving such an interpretation.  
}

Note that one choice of $\vec{Y}$ for which the condition in AC2(b) is required to hold, is $\vec{Y}=\emptyset$. For that choice, AC2 states that the effect counterfactually depends on the cause when holding fixed the witness $\vec{W}=\vec{w}$: $(M,\vec{u}) \sat  [\vec{X} \gets \vec{x}, \vec{W} \gets \vec{w}] \phi$ and $(M,\vec{u}) \sat  [\vec{X} \gets \vec{x}', \vec{W} \gets \vec{w}] \neg \phi$. Therefore AC2(a) can easily be interpeted as expressing a -- contrastive -- necessity condition: there exist contrast values $\vec{x}'$ such that if those values were to obtain, then AC2(b) no longer holds.

The problem lies with interpreting AC2(b) as expressing causal sufficiency. The main obstacle lies in the absence of the requirement that $\vec{w} = \vec{w}^*$, i.e., it is not required that the supposedly sufficient set of events $(\vec{X}=\vec{x},\vec{W}=\vec{w})$ {\em actually took place}. Therefore we cannot simply view $(\vec{X}=\vec{x},\vec{W}=\vec{w})$ itself as the causally sufficient set we are looking for. Although it cannot be excluded that the conditions imposed by invoking $\vec{Z}$ (and $\vec{Y}$) somehow ensure the existence of some other set that {\em can} be interpreted as a causally sufficient set, it is far from obvious that this is the case. This is confirmed by the fact that Halpern \& Pearl do not even offer an attempt at giving an interpretation of AC2(b) as expressing causal sufficiency.  

Matters get worse when we turn our attention to {\bf Updated HP} \citep{halpernpearl05a}:
\begin{definition}\label{def:UHP} [{\bf Updated HP}]
\begin{description}
\item[{\rm AC2(a).}] Identical to the previous one.
\item[{\rm AC2(b).}] For all subsets $\vec{V}$ of $\vec{W}$ and subsets $\vec{Y}$ of $\vec{Z} \minus \vec{X}$, we have $(M,\vec{u}) \sat  [\vec{X} \gets \vec{x}, \vec{V} \gets \vec{v}, \vec{Y} \gets \vec{y}^*] \phi$ (where $\vec{v}$ is the  restriction of $\vec{w}$ to $\vec{V}$).
\end{description}
\end{definition}

We see that AC2(b) has become even more complicated, and yet no argument is given as to how this condition formalizes causal sufficiency, despite Halpern explicitly claiming that this is what it aims to do.\footnote{Concretely, when discussing sufficient causality we find the following \citep[p. 53]{halpernbook}:

\begin{quote}
The key intuition behind the definition of sufficient causality is that not only does $\vec{X}=\vec{x}$ suffice to bring about $\phi$ in the actual context (which is the intuition that AC2(b) [from {\bf Original HP}] and AC2(b) [from {\bf Updated HP}] are trying to capture)...
\end{quote}} Instead, the updated version is justified on the basis of examples for which the previous version gave counterintuitive answers. 

As a sidenote, \cite{halpernpearl05a} also define {\em strong causation} by demanding that the following condition holds in addition to the other two:
\begin{description}
\item[{\rm AC2(c).}] For all $\vec{w} \in \R(\vec{W})$ we have that $(M,\vec{u})\sat [\vec{X} \gets \vec{x},\vec{W} \gets \vec{w}] \phi.$
\end{description}

This definition has received almost no attention in the literature, because according to Halpern \& Pearl it is too strong.\footnote{In retrospect, there is little basis for this judgment. They only discuss two examples in which strong causation diverges from {\bf Updated HP}. In the first of those (Ex. 3.2), it fails to call the lighting of each of two matches ($ML_1=1$ and $ML_2=1$) to be causes of a forest fire, whereas {\bf Updated HP} does not. However, their conjunction $(ML_1=1,ML_2=1)$ {\em is} a strong cause, and thus each of them is part of a strong cause. As we will see, Halpern later suggests treating ``part of a cause'' as being synonymous to ``cause'', so the point would be moot. In the second example (Ex. 5.5), discussed as Example \ref{trumping} later on, $S=1$ is not a strong cause although it is a cause according to {\bf Updated HP}. This is an example of {\em trumping causation}, for which the majority opinion is that $S=1$ is indeed not a cause. Moreover, Halpern's later definition {\bf Modified HP} also does not consider it a cause.} As we shall see, this is unfortunate, because AC2(c) does adequately capture a variant of causal sufficiency.

Finally we have {\bf Modified HP}, which is far simpler than the previous two \citep{halpern15}.
\begin{definition}\label{def:MHP} [{\bf Modified HP}]
\begin{description}
\item[{\rm AC2.}] There is a set $\vec{W}$ of variables in $\V \minus \vec{X}$,
and a setting $\vec{x}'$ of the variables in $\vec{X}$ such that 
$(M,\vec{u}) \sat  [\vec{X} \gets \vec{x}', \vec{W} \gets \vec{w}^* ] \neg \phi.$
\end{description}
\end{definition}

\commentout{
Here we no longer have an explicit sufficiency condition, but it is not required either: by demanding that the witness has to take on its actual values $\vec{w}^*$, both versions of AC2(b) are satisfied automatically. Halpern considers this definition to be an improvemement over the other two, and I agree with him. However, Halpern arrives at this conclusion based on the many examples in which it better agrees with intuition. As will become clear, another -- and arguably more compelling -- justification is to be found in the fact that it is the only definition of the three which {\em does} have a natural interpretation as formalizing the NESS intuition with which we started. To get there, we need to step away from the HP definitions and start afresh. 
}

The crucial difference here is that {\bf Modified HP} {\em does} require the witness to consist solely of events which actually took place, i.e., $\vec{w} = \vec{w}^*$. It is straightforward to show that simply adding this requirement ensures that both versions of AC2(b) are satisfied automatically, and therefore an explicit sufficiency condition is not required. Halpern considers this definition to be an improvemement over the other two, and I agree with him. However, Halpern arrives at this conclusion based on the many examples in which it better agrees with intuition. As will become clear, another -- and arguably more compelling -- justification is to be found in the fact that it is the only definition of the three which has a natural interpretation as formalizing the NESS intuition with which we started. To get there, we need to step away from the HP definitions and start afresh.

\section{Causal Sufficiency}\label{sec:suf}

\subsection{Some technical preliminaries}

1: \cite{halpernbook} suggests treating ``part of a cause'' (i.e., any $X=x$ that appears in $\vec{X}=\vec{x}$) as synonymous with ``cause'' when talking about {\bf Modified HP}. I will follow this suggestion throughout whenever discussing the judgment of {\bf Modified HP} in particular examples, unless stated otherwise. In stating theorems, however, the two are kept apart.

2: The HP definitions allow the effect to be any propositional formula $\phi$, whereas the other definitions of causation will require effects to be of the form $Y=y$. A thorough discussion of complex effects is beyond the scope of this paper. I here limit myself to two observations. 
\begin{itemize}
	\item Although the definitions of causation here developed can be generalized to allow for conjunctive effects (i.e., effects of the form $\vec{Y}=\vec{y}$), it is not at all clear that we should want to do so. The reason is that we can easily include variables into the effect that have nothing whatsoever to do with the causes. Say we have a variable $Y$ with equation $Y=U$, where $U$ is an exogenous variable, and we are considering a context where $U=1$. Then for any cause-effect pair $\vec{X}=\vec{x}$ and $\phi$, we automatically get that $\vec{X}=\vec{x}$ also causes $\phi \land Y=1$, which is not a sensible result. Therefore we choose to simply exclude conjunctive effects. 
	\item In the few examples in the literature where the HP definitions actually consider an effect $\phi$ that is not of the form $Y=y$, $\phi$ takes on the form $Y=y_1 \lor Y=y_2, \ldots, \lor Y=y_n$ for some $n$. The definitions here developed can easily be generalized to also allow for such effects. For reasons of simplicitly I choose not to do so in general and limit the discussion of this generalization to one example for which it is required.
\commentout{
mimic such an effect by adding a binary variable $Z$ to the causal model with equation $Z=\phi$. In this manner we can simulate $\vec{X}=\vec{x}$ causes $\phi$ by looking at $\vec{X}=\vec{x}$ causes $Z=1$. I do not want to claim that this strategy offers a general reduction of complex effects to effects of the form $Y=y$, but it does seem appropriate for these simple disjunctive cases.
}
\end{itemize}

3: The definitions of sufficiency below (and the definitions of actual causation that follow in their wake) could be extended to also allow for exogenous variables as members of a sufficient set, so that exogenous and endogenous variables are treated alike. Since our goal is to make comparisons with the HP definitions, those would also have to be extended. Concretely, the HP definitions restrict causes to being endogenous variables, and they do not allow exogenous variables to be parts of a ``witness'' (the set $\vec{W}$ above). For example, if we have $Y=X \lor U$ where $U \in \U$ and we consider a context where $U=1$ and $X=1$, the HP definitions are unable to identify $X=1$ as a cause because they disallow considering what happens when $U=0$. The simplest way to sidestep this issue is to restrict ourselves to models where exogenous variables only appear in equations of the form $V=U$. In that manner, all influence of the exogenous variables can be overriden by interventions, reducing their role to simply providing us with the actual values of all variables. For any model which does not conform to this restriction, we can easily construct a very similar model that does: simply replace any exogenous variable $U$ which appears in some equation that is not of this form with a new endogenous variable $V_U$, and add the equation $V_U=U$. For the previous example this results in the model with equations $Y=X \lor V_U$, $V_U=U$. (Note that now the HP definitions do consider $X=1$ to be a cause of $Y=1$.)

\subsection{Six Variants of Sufficiency}

Throughout the rest of the paper, we take $\vec{X}$ and $\vec{Y}$ to be non-identical subsets of the endogenous variables $\V$ that appear in a causal model $M$.\footnote{We take them to be non-identical to exclude calling a setting $\vec{X}=\vec{x}$ causally sufficient for itself, and a fortiori to exclude calling it a cause of itself.}

Informally, to say that some setting $\vec{X}=\vec{x}$ is sufficient for another setting $\vec{Y}=\vec{y}$, is to say that the latter follows from the former.\footnote{Note that in this paper we are interested in the causal sufficiency of {\em settings of variables} for other settings of variables. This is quite distinct from how the term ``causal sufficiency'' is sometimes used in the causal modelling literature, namely as a property of a {\em set of variables} in a causal graph.} To formalize this requires making explicit what it means for one setting to ``follow'' from another. In the context of {\em causal} sufficiency, an obvious minimal demand is that this meaning captures the causal directionality. In the framework of causal models this comes down to treating $\vec{X}=\vec{x}$ as an intervention and $\vec{Y}=\vec{y}$ as a consequence of that intervention: if we set $\vec{X}$ to the values $\vec{x}$, then $\vec{Y}$ takes on the values $\vec{y}$. At least this much is clear. 

Yet by saying this, we have said nothing at all about the other endogenous variables and their values, nor about the contexts in which we are evaluating the intervention. The difficulty lies in deciding what conditions we choose to impose on the other variables, both endogenous and exogenous. I consider six possible ways in which this decision can be made that are fairly natural, but this is by no means an exhaustive list.

We start with the strongest conditions possible: {\em in all contexts}, if we set $\vec{X}$ to the values $\vec{x}$, then $\vec{Y}$ takes on the values $\vec{y}$, {\em independent of the values of all other variables}.\footnote{\cite{weslake} also offers this definition of causal sufficiency to develop a definition of actual causation. He mistakenly claims that Halpern \& Pearl call this condition strong causation. As we have seen, strong causation does not require $\vec{C}$ to contain {\em all} other variables.} 

\begin{definition}\label{def:dirsuf} We say that {\em $\vec{X}=\vec{x}$ is directly sufficient for $\vec{Y}=\vec{y}$ in $M$} if for all $\vec{c} \in \R(\V - (\vec{X} \cup \vec{Y}))$ and all $\vec{u} \in \R(\U)$ we have that $(M,\vec{u})\sat [\vec{X} \gets \vec{x},\vec{C} \gets \vec{c}] \vec{Y}=\vec{y}$. 
\end{definition}

The strength of this definition is also its weakness: by putting such strong demands on the sufficient set, many interesting sets are excluded. This restrictiveness becomes apparent later on when we add a necessity condition (Proposition \ref{pro:single2}): only parents can ever be part of a minimal directly sufficient set. A trivial example illustrates this point. Say the equation for $Y$ is $Y=A$, the equation for $A$ is $A=X$, and we are looking at a context in which $X=1$.\footnote{In all examples the variables are binary unless indicated otherwise. A binary variable is a variable that has range $\{0,1\}$.} Then $X=1$ is not directly sufficient for $Y=1$, because intervening on $A$ overrides any influence of $X$ on $Y$. Still, there is clearly a sense in which $X=1$ {\em is} causally sufficient for $Y=1$. In particular, $X=1$ is directly sufficient for $(A=1,Y=1)$.

Generalizing this intuition provides us with the second form of sufficiency: there is some setting $\vec{N}=\vec{n}$ that includes $\vec{Y}=\vec{y}$, so that in all contexts, if we set $\vec{X}$ to the values $\vec{x}$, then $\vec{N}$ takes on the values $\vec{n}$, independent of the values of all other variables. This can be formulated more succinctly as: $\vec{X}=\vec{x}$ is directly sufficient for some set to which $\vec{Y}=\vec{y}$ belongs. 

\begin{definition}\label{def:strsuf} We say that {\em $\vec{X}=\vec{x}$ is strongly sufficient for $\vec{Y}=\vec{y}$ in $M$} if there exists a $\vec{N}=\vec{n}$ so that $\vec{Y} \subseteq \vec{N}$, $\vec{y}$ is the restriction of $\vec{n}$ to $\vec{Y}$, and $\vec{X}=\vec{x}$ is directly sufficient for $\vec{N}=\vec{n}$.

\end{definition}

Observe that another intuitive way of viewing $X=1$ as being causally sufficient for $Y=1$ in the simple example we just discussed, is to note that $X=1$ is directly sufficient for $A=1$ and $A=1$ is directly sufficient for $Y=1$. This intuition can also be generalized to define a form of sufficiency. Concretely, we can define {\em strong sufficiency along a network} as the transitive closure of direct sufficiency.\footnote{As with the definition of direct sufficiency, this one also appears in \cite{weslake}'s construction of actual causation, with the added requirement that $\vec{N}$ is minimal. This demand becomes redundant once we add our necessity condition. The other conditions Weslake invokes are quite complicated and do not have a counterpart in our story, which is why his definition also fails at the first strategy.}

\begin{definition}\label{def:strsuf3} We say that {\em $\vec{X}=\vec{x}$ is strongly sufficient for $\vec{Y}=\vec{y}$ in $M$ along a network $\vec{N}$} if there are (possibly overlapping) sets $\vec{N_i}$ such that $\vec{N}=\vec{Y} \cup_{i \in \{1,\ldots,k\}} \vec{N_i}$ and there exist values $\vec{n_i} \in \R(\vec{N_i})$ for each $i$ such that $\vec{X}=\vec{x}$ is directly sufficient for $\vec{N_1}=\vec{n_1}$, $\vec{N_1}=\vec{n_1}$ is directly sufficient for $\vec{N_2}=\vec{n_2}$, ..., and $\vec{N_k} = \vec{n_k}$ is directly sufficient for $\vec{Y}=\vec{y}$. 
\end{definition}

The following result shows that both forms of strong sufficiency are merely different ways of expressing the same notion of sufficiency (and hence the term is appropriately chosen). Taking in mind the earlier observation (to appear later as Proposition \ref{pro:single2}) that direct sufficiency combined with necessity is a relation between parents and children, we can safely think of a network as consisting of variables that lie on some path between $\vec{X}$ and $\vec{Y}$. Doing so will make it easier to apply the definitions of causation to examples. 

\pro\label{pro:network} $\vec{X}=\vec{x}$ is strongly sufficient for $\vec{Y}=\vec{y}$ in $M$ along a network $\vec{N}$ iff $\vec{X}=\vec{x}$ is strongly sufficient for $\vec{Y}=\vec{y}$ in $M$.
\epro

(Proofs of all Theorems are to be found in the Appendix.)

\noproofs{
\prf 
\eprf
}


Another obvious way to weaken the conditions on the values of the endogenous variables compared to direct sufficiency is to only consider the setting in which we leave the other variables alone, giving: {\em in all contexts}, if we set $\vec{X}$ to the values $\vec{x}$ {\em and do not intervene on any other variable}, then $\vec{Y}$ takes on the values $\vec{y}$.\footnote{This definition appears as just one condition in \cite{halpernbook}'s definition of {\em sufficient causality}. One of the other conditions is in fact actual causation.}

\begin{definition}\label{def:weaksuf} We say that {\em $\vec{X}=\vec{x}$ is weakly sufficient for $\vec{Y}=\vec{y}$ in $M$} if for all $\vec{u} \in \R(\U)$ we have that $(M,\vec{u})\sat [\vec{X} \gets \vec{x}] \vec{Y}=\vec{y}$. 
\end{definition}

The following straightforward result shows the relative strengths of the above three notions of sufficiency.

\pro\label{pro:relat} If $\vec{X}=\vec{x}$ is directly sufficient for $\vec{Y}=\vec{y}$ then $\vec{X}=\vec{x}$ is strongly sufficient for $\vec{Y}=\vec{y}$, and if $\vec{X}=\vec{x}$ is strongly sufficient for $\vec{Y}=\vec{y}$ then $\vec{X}=\vec{x}$ is weakly sufficient for $\vec{Y}=\vec{y}$.
\epro

So far we have considered three definitions that differ only with regards to the conditions they impose on the values of the endogenous variables: they all agreed on requiring their respective conditions to hold in all contexts. Yet questions of actual causation are posed relative to an actual context $\vec{u}$, and thus it is only natural that we should consider doing the same for questions of causal sufficiency. This adds three more definitions of sufficiency, which are simply the result of replacing the universal quantifier over contexts with a particular context that is assumed to be given. 

\begin{definition}\label{def:dirsuf2} We say that {\em $\vec{X}=\vec{x}$ is actually directly sufficient for $\vec{Y}=\vec{y}$ in $(M,\vec{u})$} if for all $\vec{c} \in \R(\V - (\vec{X} \cup \vec{Y}))$ we have that $(M,\vec{u})\sat [\vec{X} \gets \vec{x},\vec{C} \gets \vec{c}] \vec{Y}=\vec{y}$. 
\end{definition}

\begin{definition}\label{def:strsuf2} We say that {\em $\vec{X}=\vec{x}$ is actually strongly sufficient for $\vec{Y}=\vec{y}$ in $(M,\vec{u})$} if there exist $\vec{N}=\vec{n}$ so that $\vec{Y} \subseteq \vec{N}$, $\vec{y}$ is the restriction of $\vec{n}$ to $\vec{Y}$, and $\vec{X}=\vec{x}$ is actually directly sufficient for $\vec{N}=\vec{n}$.
\end{definition}

\begin{definition}\label{def:weaksuf2} We say that {\em $\vec{X}=\vec{x}$ is actually weakly sufficient for $\vec{Y}=\vec{y}$ in $(M,\vec{u})$} if $(M,\vec{u})\sat [\vec{X} \gets \vec{x}] \vec{Y}=\vec{y}$. 
\end{definition}

Obviously the counterpart of Proposition \ref{pro:relat} holds as well for these notions of actual sufficiency.

\subsection{General Form of Causal Sufficiency}

\commentout{
Now I show that all six definitions of sufficiency can be interpreted as simply putting different constraints on the parameters that occur in the following general definition of sufficiency.

\begin{definition}\label{def:gensuf} [{\bf General Definition of Sufficiency}]
We say that $\vec{X}=\vec{x}$ is sufficient for $\vec{Y}=\vec{y}$ in $M$ if there exist sets $\vec{C} \subseteq \V \minus (\vec{X} \cup \vec{Y})$, $\vec{N} \subseteq \V \minus (\vec{X} \cup \vec{Y} \cup \vec{C})$ and a setting $\vec{n} \in \R(\vec{N})$ such that for all $\vec{c} \in \R(\vec{C})$ and for all $\vec{u} \in \R(\U)$ we have that $(M,\vec{u})\sat [\vec{X} \gets \vec{x},\vec{C} \gets \vec{c}, \vec{N} \gets \vec{n}]\vec{Y}=\vec{y}$ and also for all $\vec{y}'' \in \R(\vec{Y})$ we have that $(M,\vec{u})\sat [\vec{X} \gets \vec{x},\vec{C} \gets \vec{c}, \vec{Y} \gets \vec{y}''] \vec{N}=\vec{n}$.

We say that $\vec{X}=\vec{x}$ is sufficient for $\vec{Y}=\vec{y}$ in $M$ along $\vec{N}$ independent of $\vec{C}$. 
\end{definition}

Although this general definition is more complicated when compared to each of the three variants of sufficiency separately, it allows us to better understand the connection between them.

\pro\label{pro:suf}
The three variants of sufficiency are equivalent to Definition \ref{def:gensuf} when making the following choices for $\vec{N}$ and $\vec{C}$:
\begin{description}
\item[{\rm Weak Sufficiency.}] Choose both $\vec{C}$ and $\vec{N}$ to be minimal, i.e., $\vec{C}=\vec{N}=\emptyset$.
\item[{\rm Strong Sufficiency.}] Choose $\vec{N}$ to be maximal given $\vec{C}$, i.e., $\vec{N}=\V \minus (\vec{X} \cup \vec{Y} \cup \vec{C})$.
\item[{\rm Direct Sufficiency.}] Choose $\vec{C}$ to be maximal, i.e., $\vec{C} = \V \minus (\vec{X} \cup \vec{Y})$ and thus $\vec{N}=\emptyset$.
\end{description}
\epro

\noproofs{
\prf 
\eprf
}

Proposition \ref{pro:suf} could inspire even more variants of sufficiency. In fact, we have already come across the most obvious one: AC2(c). It is easy to see that it consists of choosing $\vec{N}$ to be minimal given $\vec{C}$, i.e., $\vec{N}=\emptyset$, meaning it sits in between Weak and Strong Sufficiency. The condition also appears as a sufficiency condition in Pearl's notion of {\em sustenance}, which is the first step he takes towards formalizing the NESS intuition (\citeyear[p. 317]{pearl:book2}). Unfortunately it is also the last step, because the subsequent notions he introduces are far more complicated and bear no resemblance to NESS. The added complexity is introduced precisely because taken by itself sustenance fails to provide a sensible definition of causation, which is why I leave the exploration of this and other possible variants of sufficiency for another occasion.

We can also define {\em actual sufficiency} by holding fixed the actual context in Definition \ref{def:gensuf}. Therefore strictly speaking we have six definitions of sufficiency in total, but as we remarked earlier, the distinction between actual and non-actual sufficiency only matters for Weak Sufficiency. So we only have four definitions of causal sufficiency that are interesting.

Given that Definition \ref{def:gensuf} is needlessly complicated when compared to Definitions \ref{def:dirsuf} and \ref{def:weaksuf}, it is easier to use the latter when working with Direct and Weak Sufficiency. For Strong Sufficiency, Definition \ref{def:gensuf} can be made slightly simpler once we restrict ourselves to cases where $\vec{Y}$ is a single conjunct:

\pro\label{pro:sim} 
$\vec{X}=\vec{x}$ is strongly sufficient for $Y=y$ iff there exists some $\vec{N} \subseteq (\V \minus (\vec{X} \cup \{Y\}))$ and a setting $\vec{n} \in \R(\vec{N})$ such that for all $\vec{c} \in \R(\V \minus (\vec{N} \cup \vec{X} \cup \{Y\}))$ and for all $\vec{u} \in \R(\U)$ we have that $(M,\vec{u})\sat [\vec{X} \gets \vec{x},\vec{C} \gets \vec{c}, \vec{N} \gets \vec{n}]Y=y$ and $(M,\vec{u})\sat [\vec{X} \gets \vec{x},\vec{C} \gets \vec{c}] \vec{N}=\vec{n}$.
\epro

\noproofs{
\prf 
\eprf
}
}

We can formalize and generalize the intuitions behind the definitions in the preceding section by showing that all six definitions of sufficiency can be interpreted as simply putting different constraints on the parameters that occur in the following general definition of sufficiency. (We only explicitly discuss the three definitions of ``non-actual'' sufficiency, but the same analysis trivially applies to the three definitions of actual sufficiency.)
\begin{definition}\label{def:gensuf} [{\bf General Definition of Sufficiency}]
We say that $\vec{X}=\vec{x}$ is {\em sufficient} for $\vec{Y}=\vec{y}$ in $M$ if there exist sets $\vec{C} \subseteq \V \minus (\vec{X} \cup \vec{Y})$, $\vec{N} \subseteq \V \minus (\vec{X} \cup \vec{C})$ with $\vec{Y} \subseteq \vec{N}$, and a setting $\vec{n} \in \R(\vec{N})$ where the restriction of $\vec{n}$ to $\vec{Y}$ is $\vec{y}$, such that for all $\vec{c} \in \R(\vec{C})$ and for all $\vec{u} \in \R(\U)$ we have that $(M,\vec{u})\sat [\vec{X} \gets \vec{x},\vec{C} \gets \vec{c}]\vec{N}=\vec{n}$.

We say that $\vec{X}=\vec{x}$ is sufficient for $\vec{Y}=\vec{y}$ in $M$ along $\vec{N}$ independent of $\vec{C}$. 
\end{definition}

This definition is more complicated than Definitions \ref{def:dirsuf}, \ref{def:strsuf}, and \ref{def:weaksuf}. Its use lies in the fact that it allows us to see exactly how the three definitions relate to each other, and how one can construct other definitions of sufficiency, by invoking the following trivial result.

\pro\label{pro:suf}
Definitions \ref{def:dirsuf}, \ref{def:strsuf}, and \ref{def:weaksuf}, are equivalent to Definition \ref{def:gensuf} when making respectively the following choices for $\vec{N}$ and $\vec{C}$:
\begin{description}
\item[{\rm Weak Sufficiency.}] Choose both $\vec{C}$ and $\vec{N}$ to be minimal, i.e., $\vec{C}=\emptyset$, $\vec{N}=\vec{Y}$.
\item[{\rm Strong Sufficiency.}] Choose $\vec{N}$ to be maximal given $\vec{C}$, i.e., $\vec{N}=\V \minus (\vec{X} \cup \vec{C})$.
\item[{\rm Direct Sufficiency.}] Choose $\vec{C}$ to be maximal, i.e., $\vec{C} = \V \minus (\vec{X} \cup \vec{Y})$ and thus $\vec{N}=\vec{Y}$.
\end{description}
\epro

Proposition \ref{pro:suf} could inspire even more variants of sufficiency. In fact, we have already come across the most obvious one: AC2(c). It is easy to see that it consists of choosing $\vec{N}$ to be minimal given $\vec{C}$, i.e., 
$\vec{N}=\vec{Y}$,
 meaning it sits in between Weak and Strong Sufficiency. The condition also appears as a sufficiency condition in Pearl's notion of {\em sustenance}, which is the first step he takes towards formalizing the NESS intuition (\citeyear[p. 317]{pearl:book2}). Unfortunately it is also the last step, because the subsequent notions he introduces are far more complicated and bear no resemblance to NESS. The added complexity is introduced precisely because taken by itself sustenance fails to provide a sensible definition of causation, which is why I leave the exploration of this and other possible variants of sufficiency for another occasion.

\section{Defining Causation using Sufficiency}\label{sec:ac}

We are finally ready to take up the main challenge: defining actual causation as the formal expression of the NESS intuition. In order to do so, several questions need  to be answered: 
\begin{itemize}
	\item 
	Should we use actual sufficiency or not?
	\item Which of the three definitions of (actual) causal sufficiency should we use?
	\item Does necessity mean that there exist contrast values of $\vec{X}$ so that the set would not be sufficient if those values obtained, or does it mean that the set is no longer sufficient when we remove the subset $\vec{X}$?
\end{itemize}

I have introduced six definitions of causal sufficiency in the previous section. For each definition, we can define causation using either of the two interpretations of necessity, giving twelve definitions of actual causation altogether. However, I will show that several of these are equivalent to each other, and one will be impossible to satisfy, leaving us with six definitions in the end. One of those will be {\bf Modified HP}.

\subsection{A Family of Definitions}

As with the HP definitions, Definition \ref{def:HP} gives the general form of all definitions, except that $\phi$ is restricted to $Y=y$. (This restriction is assumed whenever comparisons are made with the HP definitions.) As before, the only difference lies with the content of AC2. Using the first interpretation of necessity, which we shall call {\em contrastive necessity}, the general form of AC2 is as follows:

\begin{definition}\label{def:gen} [{\bf General Definition of Causation}]
There exist sets 
$\vec{W},\vec{N}$ such that
\begin{description}
\item[{\rm AC2($\text{a}^{\text{c}}$).}] There exist values $\vec{x}'$ such that for all $\vec{S} \subseteq \vec{N}$, $(\vec{X}=\vec{x}',\vec{W}=\vec{w}^*)$ is not sufficient for $Y=y$ along $\vec{S}$.
\item[{\rm AC2(b).}] $(\vec{X}=\vec{x},\vec{W}=\vec{w}^*)$ is sufficient for $Y=y$ along $\vec{N}$.
\end{description}
We call $\vec{W}$ a {\em witness} of $\vec{X}=\vec{x}$ causing $Y=y$.
\end{definition}
\commentout{
Can we change network to path? No. Ex.

Y = C&D&A&B v not-X&A v not-X&B
C = A
D = B
A = X
B = X
}
\commentout{
Can we always take W to be everything except for N? No, of course not: Y = X v W. But if you demand it nonetheless, and add that N is a path, do you get Weslake's pre? Don't know, not interesting enough.
}

By replacing sufficiency in the {\bf General Definition of Causation} with any of the six definitions of sufficiency from Section \ref{sec:suf}, we obtain six specific definitions of actual causation.\footnote{Definition \ref{def:gen} can be made even more general by also incorporating $\vec{C}$ from Definition \ref{def:gensuf}. Since we are only considering notions of sufficiency for which $\vec{C}$ is determined entirely by the other sets, there is no need to do so for our purposes. But it is important to keep this additional generality in mind if one wants to use alternative definitions of sufficiency.} AC2(b) simply expresses causal sufficiency, whatever form it may take. AC2($\text{a}^{\text{c}}$) offers a somewhat nuanced expression of necessity because it also focusses on subsets of $\vec{N}$. (Note that this nuance matters only for Strong Sufficiency, since for Weak and Direct Sufficiency $\vec{N}=\{Y\}$ anyway.) 
\commentout{The reason is that our interest lies with the sufficiency for $Y=y$, and the network $\vec{N}$ is merely a means to an end. So if $(\vec{X}=\vec{x}',\vec{W}=\vec{w}^*)$ is able to achieve sufficiency with less means than $(\vec{X}=\vec{x},\vec{W}=\vec{w}^*)$, $\vec{X}=\vec{x}$ is not necessary for the sufficiency of the latter. Informally, ``along a network'' is interpreted in AC2(a) as meaning ``along some variables within a network'', rather than ``along all variables within a network''.
}
The reason is that our interest lies with the sufficiency for $Y=y$, and the network $\vec{N}$ is merely a means to that end. If $\vec{X}=\vec{x}'$ accomplishes the same end using less means, then $\vec{X}=\vec{x}$ was not necessary for achieving it.

Under the second interpretation of necessity, which we shall call {\em minimal necessity}, AC2($\text{a}^{\text{c}}$) is replaced with:
\begin{description}
\item[{\rm AC2($\text{a}^{\text{m}}$).}] For all $\vec{S} \subseteq \vec{N}$, $\vec{W}=\vec{w}^*$ is not sufficient for $Y=y$ along $\vec{S}$.
\end{description}

Both interpretations of necessity are prima facie plausible. The contrastive interpretation is explicitly counterfactual in nature, whereas the minimal interpretation is more neutral. Our analysis will settle which one of them is to be preferred.

Filling in each of the six definitions of causal sufficiency into both versions of the {\bf General Definition of Causation} gives twelve specific definitions of actual causation. I refer to each of these as {\bf Def $x$} for $x \in \{1,\ldots,12\}$ along the following convention: 
\commentout{
\begin{itemize}
	\item {\bf Def 1} Contrastive actual weak sufficiency
	\item {\bf Def 2} Contrastive weak sufficiency
	\item {\bf Def 3} Contrastive strong sufficiency
	\item {\bf Def 4} Contrastive direct sufficiency
	\item {\bf Def 5} Minimal actual weak sufficiency
	\item {\bf Def 6} Minimal weak sufficiency
	\item {\bf Def 7} Minimal strong sufficiency
	\item {\bf Def 8} Minimal direct sufficiency
\end{itemize}
}
\begin{itemize}
	\item {\bf Def 1} Contrastive actual weak sufficiency
	\item {\bf Def 2} Contrastive actual strong sufficiency
	\item {\bf Def 3} Contrastive actual direct sufficiency
	\item {\bf Def 4} Contrastive weak sufficiency
	\item {\bf Def 5} Contrastive strong sufficiency
	\item {\bf Def 6} Contrastive direct sufficiency
	\item {\bf Def 7} Minimal actual weak sufficiency
	\item {\bf Def 8} Minimal actual strong sufficiency
	\item {\bf Def 9} Minimal actual direct sufficiency
	\item {\bf Def 10} Minimal weak sufficiency
	\item {\bf Def 11} Minimal strong sufficiency
	\item {\bf Def 12} Minimal direct sufficiency
\end{itemize}

So to be clear, each {\bf Def $x$} is constructed by taking the respective definition of sufficiency (i.e., Definition \ref{def:dirsuf}, \ref{def:strsuf}, \ref{def:weaksuf}, \ref{def:dirsuf2}, \ref{def:strsuf2}, or \ref{def:weaksuf2}), filling that into the {\bf General Definition of Causation} where AC2(a) takes on AC2($\text{a}^{\text{c}}$) or AC2($\text{a}^{\text{m}}$) depending on whether $x < 7$ or not, and finally, filling those conditions AC2 into Definition \ref{def:HP}. I illustrate the result of this construction for {\bf Def 2}.

\begin{definition}\label{def:def2}[{\bf Def 2}]
$\vec{X} = \vec{x}$ is an \emph{actual cause} of $Y=y$ according to {\bf Def 2} in $(M,\vec{u})$  if the following three conditions hold: 
\begin{description}
\item[{\rm AC1.}] $(M,\vec{u}) \sat (\vec{X} =\vec{x}) \land Y=y$. 
\item[{\rm AC2($\text{a}^{\text{c}}$).}] There exist sets $\vec{W}$, $\vec{N}$ with $Y \in \vec{N}$, and values $\vec{x}'$, such that for all $\vec{S} \subseteq \vec{N}$ with $Y \in \vec{S}$, and for all $\vec{s} \in \R(\vec{S})$ such that $y \in \vec{s}$, there exists a $\vec{t} \in \R(\V \minus (\vec{X} \cup \vec{W} \cup \vec{S}))$ so that $(M,\vec{u})\sat [\vec{X} \gets \vec{x}',\vec{W} \gets \vec{w}^*,\vec{T} \gets \vec{t}] \vec{S} \neq \vec{s}$. 
\item[{\rm AC2(b).}] For all $\vec{c} \in \R(\V - (\vec{X} \cup \vec{W} \cup \vec{N}))$ we have that $(M,\vec{u})\sat [\vec{X} \gets \vec{x}, \vec{W} \gets \vec{w}^*, \vec{C} \gets \vec{c}] \vec{N}=\vec{n}^*$. 
\item[{\rm AC3.}] $\vec{X}$ is minimal.
\end{description}
\end{definition}

Admittedly, {\bf Def 2} looks even more complicated than {\bf Updated HP}. Further on I provide some results that allow us in many cases to use simpler definitions as stand-ins for {\bf Def 2}. More importantly, although the notation of Definition \ref{def:def2} is complicated, its meaning can be spelled out intuitively by stating that $\vec{X}=\vec{x}$ causes $Y=y$ iff $\vec{X}=\vec{x}$ is a Minimal Contrastively Necessary Subset of a Strongly Sufficient Set for $Y=y$ (or  $\text{MCNS}^4$).\footnote{Strictly speaking it should say ``Actually Strongly Sufficient'', but that makes for a less elegant acronym. I am cheating a bit by anticipating Theorem \ref{thm:ident}.}

\subsection{Analysis}

Let us now turn to investigating the relations between these definitions. (Knowing these relations before getting into the discussion of examples makes life a lot easier.) A first remark is that {\bf Def 7} is impossible to satisfy, as it requires that both $(M,\vec{u}) \sat [\vec{X} \gets \vec{x}^*, \vec{W} \gets \vec{w}^*]Y=y$ and $(M,\vec{u}) \not \sat [\vec{W} \gets \vec{w}^*]Y=y$ hold, implying that $(M,\vec{u}) \sat Y=y \land  Y \neq y$. 

A second remark is that {\bf Def 3} is equivalent to a condition that appears in Pearl's first definition of actual causation (\citeyear{pearl98}).\footnote{It re-appears in his second definition of actual causation in the notion of a {\em causal beam}, but without the necessity condition \citep[p. 318]{pearl:book2}. To see the equivalence, one needs to invoke Proposition \ref{pro:single2}.} 

Ignoring {\bf Def 7}, we are still left with eleven candidate definitions of actual causation (fourteen candidates if we count the three HP definitions), whereas we would like to settle on just one. The rest of the paper is concerned with selecting the best definition out of the lot. As a first step, we can reduce the number of definitions by six.

\thm\label{thm:ident} The following are all equivalences among the twelve definitions and the three HP definitions:
\begin{itemize}
	\item  {\bf Modified HP} iff {\bf Def 1}
	\item  {\bf Def 2} iff {\bf Def 5}
	\item  {\bf Def 8} iff {\bf Def 11}
	\item  {\bf Def 3} iff {\bf Def 6} iff {\bf Def 9} iff {\bf Def 12}
\end{itemize}
\ethm 


\noproofs{
\prf 
\eprf
}

Theorem \ref{thm:ident} offers our first interesting result: it shows that {\bf Modified HP} succeeds in formalizing the NESS intuition, whereas the other two HP definitions do not. From now on I will ignore the definitions appearing on the right-hand side in Theorem \ref{thm:ident}. The following is a helpful result for applying some of the definitions going forward. (As is well known, the same result holds for {\bf Original HP} \citep{halpernbook}.)

\pro\label{pro:single} If $\vec{X}=\vec{x}$ causes $Y=y$ in $(M,\vec{u})$ according to a definition that uses minimal necessity, then $\vec{X}$ is a singleton.
\epro

\noproofs{
\prf 
\eprf
}

The following result offers important insights into the relations between the remaining definitions. 

\thm\label{thm:relations} The only implications -- involving either causes or parts of causes -- between the remaining five definitions ({\bf Def 2}, {\bf Def 3}, {\bf Def 4}, {\bf Def 8}, and {\bf Def 10}) and the three HP definitions are the following ones (and their immediate consequences, of course):	
\begin{itemize}
	\item If part of {\bf Modified HP} then {\bf Updated HP};\footnote{This is shorthand for: If $X=x$ is part of a cause of $Y=y$ according to the {\bf Modified HP} definition then it is a cause of $Y=y$ according to the {\bf Updated HP} definition.}
	\item If part of {\bf Updated HP} then {\bf Original HP}; 
	\item If {\bf Def 3} then {\bf Def 2};
	\item If part of {\bf Def 2} then {\bf Def 8};
	\item If {\bf Def 3} then {\bf Original HP};
	\item If {\bf Def 10} then {\bf Def 4}. 
\end{itemize}
\ethm

\noproofs{
\prf 
\eprf
}

\section{Excluding {\bf Def 3} and {\bf Def 10}}\label{sec:4and6}

Two definitions can be excluded quickly. The following result shows why {\bf Def 3} is not a sensible candidate as a general definition of causation, since causation is obviously not restricted to parent-children pairs.

\pro\label{pro:single2} If $\vec{X}=\vec{x}$ causes $Y=y$ in $(M,\vec{u})$ according to {\bf Def 3}, then $\vec{X}$ is a singleton, and $X$ is a parent of $Y$. 
\epro

\noproofs{
\prf 
\eprf
}

Although we can dismiss {\bf Def 3} as a general definition of causation, it is still a useful stand-in for -- the arguably more complicated -- {\bf Def 2} and {\bf Def 8} in case $X$ is a parent of $Y$ and $X$ is not an ancestor of $Y$ along any path that is longer than a single edge (which in fact covers a surprisingly large number of cases discussed in the literature). In such cases we say that $X$ is {\em only} a parent of $Y$.

\pro\label{pro:parent} If $X$ is only a parent of $Y$, then {\bf Def 2}, {\bf Def 3}, and {\bf Def 8} are all equivalent for causes $X=x$.
\epro

\noproofs{
\prf 
\eprf
}

A cornerstone of the counterfactual approach to causation is that counterfactual dependence is sufficient for causation. More formally, there is widespread consensus that causation should satisfy the following principle:\footnote{As does Halpern, I here restrict myself to counterfactual dependence on a single conjunct (\citeyear[p. 26]{halpernbook}).}

\begin{principle}[{\bf Dependence}] Say $(M,\vec{u}) \sat X=x \land Y=y$. If there exists a value $x'$ such that $(M,\vec{u}) \sat [X \gets x']Y \neq y$ then $X=x$ causes $Y=y$ in $(M,\vec{u})$.
\end{principle}

Accepting this principle means that {\bf Def 10} is excluded as well. 

\pro\label{pro:dep} Out of all definitions we have considered, {\bf Def 10} and {\bf Def 3} are the only ones which do not satisfy {\bf Dependence}.
\epro 

\noproofs{
\prf
\eprf
}

That leaves us with {\bf Def 2}, {\bf Def 4}, and {\bf Def 8} as possible alternatives to the HP definitions.

\section{{\bf Def 2}, {\bf Def 4}, and {\bf Def 8}, vs the HP definitions}\label{sec:237}

We have shown that all twelve definitions we developed (including {\bf Modified HP}) are instantiations of the {\bf General Definition of Causation} (Def. \ref{def:gen}), and thereby they improve upon {\bf Original HP} and {\bf Updated HP} as far as the first strategy goes. We now show that {\bf Def 2} also improves upon all three HP definitions as far as the second strategy goes, whereas {\bf Def 4} and {\bf Def 8} do not. In order to remain as neutral as possible, we go over Halpern \& Pearl's own examples, compare the verdicts of our definitions to theirs, and stick as close as possible to their intuitions.

\subsection{Comparison to {\bf Updated HP}}

The {\bf Updated HP} definition is by far the most well-known. It was developed as an improvement of {\bf Original HP}, which sometimes gives unreasonable answers.  \cite{halpernpearl05a} offer many examples to illustrate how it works and how it successfully deals with paradigm cases of causation. 

Their first example is one of those few cases -- recall the beginning of Section \ref{sec:suf} -- in which the effect is of the form $Y=y_1 \lor Y=y_2$, and therefore allows us to illustrate how we can generalize the {\bf General Definition of Causation} to such effects. It is also an example for which {\bf Def 8} gives the wrong answer, but the subsequent example is far simpler and more convincing in this respect.

\begin{example}\label{storm} ``Suppose that there was a heavy rain in April and electrical storms in the following two months; and in June the lightning took hold. If it hadn't been for the heavy rain in April, the forest would have caught fire in May.'' \citep[p. 15]{halpernpearl05a} I agree with Halpern and Pearl's judgment that it would be very counterintuitive to say that the April rain caused the forest fire, since all it did was delay the fire. As they indicate, it is nevertheless perfectly sensible to say that the April rain caused the forest fire {\em in June}, as opposed to May. In order to capture this distinction, we need to invoke a disjunctive effect. 
	
Let $F$ represent there being a fire or not, with three possible values: $0$ (no fire), $1$ (fire in May), or $2$ (fire in June). $ES$ is a four-valued variable that captures whether there are electric storms: $(0,0)$ (no electric storms in either May or June), $(1,0)$ (electric storms in May but not in June), $(0,1)$ (storms in June but not May), and $(1,1)$ (storms in both May and June). Lastly, $AS$ is a binary variable expressing whether or not there was April rain.

The equation for $F$ is then given by: $F=2$ if $(AS=1 \land ES=(1,1)) \lor ES=(0,1)$, $F=1$ if $AS=0 \land (ES=(1,1) \lor ES=(1,0))$, and $F=0$ otherwise. Given that $F=2$ counterfactually depends on $AS=1$, all definitions we are considering agree that $AS=1$ causes $F=2$. The question is whether $AS=1$ also caused there to be a fire, i.e., whether it caused $F=1 \lor F=2$. 

We can easily generalize sufficiency to such disjunctions: $\vec{X}=\vec{x}$ is sufficient for $Y=y \lor Y=y'$ iff $\vec{X}=\vec{x}$ is sufficient for $Y=y$ or $\vec{X}=\vec{x}$ is sufficient for $Y=y'$. When integrated into our {\bf General Definition of Causation}, this results in splitting up AC2(a) so that there is one instance for each disjunct. AC2(b) need not be split up, since it can only ever be satisfied for the actual value of $Y$.\footnote{Note that this means generalizing to disjunctions across different variables -- i.e., something like $Y=y \lor Z=z$ -- is more complicated.} 

Let us apply this idea to our example. To satisfy AC2(b), we have to add $ES$ to the witness: $(AS=1,ES=(1,1))$ is directly sufficient for $F=2$ and $AS=1$ is not. (We can focus on direct sufficiency because $AS$ is only a parent of $F$. We cannot invoke Proposition \ref{pro:parent} though, since that requires an effect $Y=y$.) We then see that one of the two conditions that now make up AC2(a) is not satisfied for {\bf Def 2} and {\bf Def 4}, because $(AS=0,ES=(1,1))$ is directly sufficient for $F=1$. Therefore {\bf Def 2} and {\bf Def 4} agree with the HP definitions that the April rain did not cause the forest fire. But {\bf Def 8} does not reach this verdict, because $ES=(1,1)$ is not directly sufficient for either $F=1$, nor is it for $F=2$. This means AC2(a) is fullfilled for {\bf Def 8}, which leads to a mistaken conclusion. 
\end{example}

Although one counterexample need not disqualify a definition, the following example is indicative of a deeper problem with {\bf Def 8}: whenever $X=x$ strongly suffices for $Y=y$, it is automatically a cause according to {\bf Def 8}, since $\emptyset$ is never strongly sufficient for $Y=y$. The following example is but one of many paradigm cases in the literature for which this property leads to a counterintuitive verdict.\footnote{\cite{mcdermott95} offers an almost identical example involving a dog biting a terrorist. Another famous case is that involving a boulder rolling towards a hiker \citep{hitchcock01}. All of these examples are counterexamples to the transitivity of causation. The failure of transitivity has become broadly accepted by now \citep{beckers17}. Despite what {\bf Def 8}'s behavior in these examples might suggest, it is also not transitive. A simple counterexample consists of equations $Z=Y \lor W$, and $Y=X\land W$. If $X=W=1$, {\bf Def 8} considers $X=1$ a cause of $Y=1$, $Y=1$ a cause of $Z=1$, yet it does not consider $X=1$ a cause of $Z=1$.} Therefore {\bf Def 8} is also excluded as a definition of causation.

\begin{example}\label{switch} ``The engineer is standing by a switch in the railroad tracks. A train approaches in the distance. She flips the switch, so that the train travels down the right-hand track, instead of the left. Since the tracks reconverge up ahead, the train arrives at its destination all the same...
	
Again, our causal model gets this right. Suppose we have three random variables:
\begin{itemize}
\item $F$ for ``flip'', with values $0$ (the engineer doesn’t flip the switch) and $1$ (she does);
\item $T$ for ``track'', with values $0$ (the train goes on the left-hand track) and $1$ (it goes on the right-hand track); and
\item $A$ for ``arrival'', with values $0$ (the train does not arrive at the point of reconvergence) and $1$ (it does).
\end{itemize}'' \citep[p. 26]{halpernpearl05a}

First observe that as described, this causal model makes little sense: the equation for $A$ is given by $A=T \lor \lnot T$, which can be rewritten as $A=1$. This can be fixed by extending the range of $T$ with a value $2$, representing the train not going down any track (because it breaks down, for example). Then the equations become $A= (T \neq 2)$ and $T=F$. The context is such that $F=1$. 

$F=1$ is both weakly sufficient for $A=1$ and strongly sufficient for $A=1$ along $\{T\}$, but so is $F=0$. Therefore {\bf Def 2} and {\bf Def 4} agree with {\bf Updated HP} (and with intuition) that flipping the switch is not a cause of the train's arrival. {\bf Def 8} fails to reach this verdict, because $\emptyset$ is not strongly sufficient for $A=1$.
\end{example}


{\bf Def 4} suffers from an even bigger defect than {\bf Def 8}: it fails to distinguish preempted causes from preempting causes. Since preemption cases are the bread and butter of the literature on actual causation, this means that {\bf Def 4} is immediately disqualified. The following is a famous example of late preemption discussed by \cite{halpernpearl05a} (and originally by \cite{hall04}).

\begin{example}\label{LP}
Suzy and Billy both throw a rock at a bottle. Suzy's rock gets there first, shattering the bottle. However Billy's throw was also accurate, and would have shattered the bottle had it not been preempted by Suzy's throw. \cite{halpernpearl05a} use the following variables for this example, which capture the fact that Billy's throw was preempted by Suzy's rock hitting the bottle: $BS$ for the bottle shattering, $BH$, $SH$ for Billy's (resp. Suzy's) rock hitting the bottle, and two more variables ($BT$, $ST$) for either of them throwing their rock. The equations are then as follows: $BS=BH \lor SH$, $SH=ST$, $BH=BT \land \lnot SH$. None of the definitions has any problem arriving at the obvious result that Suzy's throw ($ST=1$) causes the bottle to shatter ($BS=1$). However, {\bf Def 4} is the only definition under consideration that mistakenly also judges Billy's throw to be a cause of  the bottle's shattering: in all contexts $BT=1$ is weakly sufficient for $BS=1$, whereas $BT=0$ is not weakly sufficient for $BS=1$ in the context where $ST=0$.
\end{example}

This leaves us with {\bf Def 2} as the last potential alternative to the HP definitions. Going through the many remaining examples, there is only one in which {\bf Def 2} disagrees with {\bf Updated HP}. I leave it to the reader to verify this claim, and restrict the discussion to that single example.

\begin{example}\label{trumping} Major ($M$) and sergeant ($S$) stand before corporal, and both shout `Charge!' ($M = 1$, $S = 1$). The corporal charges ($C = 1$). Orders from higher-ranking soldiers trump those of lower rank, so if the major had shouted `Halt' ($M = 0$) the corporal would not have charged. If the major remains quiet ($M=-1$), the corporal listens to the sergeant.\footnote{This formulation is due to \cite{weslake}, but the example was first discussed by \cite{schaffer00} (who attributes it to van Fraassen).} The equation for $C$ is thus: $C= M$ if $M \neq -1$ and $C=S$ otherwise. The majority intuition is that the sergeant did not cause the corporal to charge, because his order was trumped by that of the major.\footnote{See \cite{weslake} for a discussion.}
	
{\bf Def 2} agrees, as it does not consider $S=1$ a cause of $C=1$. The reason is that $M=1$ is directly sufficient by itself, and yet $S=1$ needs $M=1$ as a witness to form a sufficient set. $S=1$ is a cause of $C=1$ according to both {\bf Original HP} and {\bf Updated HP}. Halpern \& Pearl do not consider this to be problematic, but they do go through the trouble of showing how {\bf Original HP} and {\bf Updated HP} change their verdict if one adds extra variables to the model. Moreover, {\bf Modified HP} also agrees with {\bf Def 2} here. Given Halpern's later preference for {\bf Modified HP}, it is fair to say that {\bf Def 2} does at least as good as {\bf Updated HP} on this example.
\end{example}

\subsection{Comparison to {\bf Modified HP}}\label{sec:modhp}

Dissatisfied with {\bf Updated HP} due to the many counterexamples that were presented in the literature, \cite{halpern15} develops {\bf Modified HP}. First of all, despite Theorem \ref{thm:relations}, there do exist interesting connections between the three definitions we have considered and {\bf Modified HP}.

\pro\label{pro:mod} If {\bf Modified HP} with $\vec{X}$ a singleton, then {\bf Def 2}, {\bf Def 4}, and {\bf Def 8}.
\epro

\noproofs{
\prf
\eprf
}

\cite{halpern15} goes over several counterexamples to {\bf Updated HP} and shows that {\bf Modified HP} offers sensible verdicts. Taking into account Halpern's suggestion that ``part of cause'' is synonymous with ``cause'' for {\bf Modified HP}, there are in fact only three examples in which {\bf Modified HP} disagrees with {\bf Updated HP} (Examples 3.5, 3.8, and 3.11).\footnote{When discussing Example 3.8 again in \citep{halpernbook}, he mistakenly claims that {\bf Modified HP} agrees with {\bf Updated HP} when treating parts of causes as causes. In response, Halpern has suggested a small variation on the example in which {\bf Modified HP} indeed does agree with {\bf Updated HP} (personal communication). For that variation, {\bf Def 2} also agrees with the HP definitions.} In all three of those cases, {\bf Def 2} sides with {\bf Modified HP}.

There is only one example in which {\bf Def 2} disagrees with {\bf Modified HP}.\footnote{\cite{halpernbook} discusses far more cases, but none of them reveal any further disagreements between these definitions.} Crucially, it is an example for which Halpern agrees that {\bf Modified HP} reaches the wrong verdict. 

\begin{example}\label{voting} A ranch has five individuals: $a_1,\ldots,a_5$. They have to vote on two possible outcomes: staying at the campfire ($O=0$) or going on a round-up ($O=1$). Let $A_i$ be the random variable denoting $a_i$'s vote, so $A_i = j$ if $a_i$ votes for outcome $j$. There is a complicated rule for deciding on the outcome. If $a_1$ and $a_2$ agree (i.e., if $A_1 = A_2$), then that is the outcome. If $a_2,\ldots,a_5$ agree, and $a_1$ votes differently, then the outcome is given by $a_1$'s vote (i.e., $O = A_1$). Otherwise, majority rules. In the actual situation, $A_1 =A_2 =1$ and $A_3 =A_4 =A_5 =0$, so by the first mechanism, $O = 1$.\footnote{This is the formulation of the example found in \citep[p. 109]{halpernbook}, but the example was first presented by \cite{stonesoup}.}

Halpern states, and I agree, that intuitively one should expect only $A_1=1$ and $A_2=1$ to be causes of $O = 1$. After all, $a_3,\ldots,a_5$ voted {\em against} $O=1$. {\bf Def 2} gives that result, whereas {\bf Modified HP} considers every vote to be a cause. Halpern argues for adding more variables to the model in order to get the right outcome, but it speaks in favor of {\bf Def 2} that it is able to give the right answer with just these variables.
\end{example}

We conclude that judged by the second strategy and Halpern \& Pearl's own examples, {\bf Def 2} does better than {\bf Updated HP} and at least as good as {\bf Modified HP}. Lastly we consider a very simple example that was offered as a counterexample to {\bf Modified HP} by \cite{halpern_review}.

\begin{example}\label{ex:noname} 
We have equations $Y=X \lor D$ and $X=D$, and we consider a context such that $D=1$. This looks very much like a standard case of overdetermination in which $X=1$ and $D=1$ are both overdetermining causes. That is also the verdict of all of the definitions considered in this paper, except for {\bf Modified HP}: it does not consider $X=1$ a cause of $Y=1$. The reason for this is that $Y=1$ depends counterfactually on $D=1$ by itself, whereas it does not depend on $X=1$ by itself and nor does it when we take $D=1$ as a witness. \cite{halpern_review} state that Halpern endorses this conclusion, but offer the following story to motivate why they consider that an untenable position.

``An obedient gang is ordered by its leader to join him in murdering someone, and does so, all of them shooting the victim at the same time, or all of them together pushing the plunger connected to a bomb. The action of any one of the gang would suffice for the victim's death. If responsibility implies causality, whom among them is responsible? Were you among the jury, whom would you convict? What ought the Hague Court to do in cases of subordinates sure to obey orders? Halpern's theory says the gang leader and only the gang leader is a cause of the victim's death. This is a morally intolerable result; absent a plausible general principle severing responsibility from causation, any theory that yields such a result should be rejected.''  
\end{example}

Even if one disagrees with this judgment, the next section offers further motivation for preferring {\bf Def 2} over {\bf Modified HP}.

\subsection{{\bf Def 2} vs the Others}

Finally I will argue that {\bf Def 2} does better than all of the other definitions on a few more examples according to two metrics: it offers verdicts that are both intuitively plausible {\em and} consistent across minor changes of the examples. Before doing so, I present an example that illustrates a special property of {\bf Def 2}. 

Recall from Section \ref{sec:hp} that it is a necessary condition for all three HP definitions that there exists some  $[\vec{W} \gets \vec{w}]$ such that $Y=y$ counterfactually depends on $\vec{X}=\vec{x}$ under that intervention. The same is true for the most well-known definitions out there that have been inspired by the HP definitions (see \cite{weslake} for an overview), as well as for {\bf Def 3}, {\bf Def 4}, and {\bf Def 10}. Let us call definitions with this property {\em strongly counterfactual}. Although {\bf Def 2} clearly also relies on counterfactuals, and thus falls within the counterfactual approach to causation, it is not strongly counterfactual, as the following example shows.\footnote{It is not so clear that {\bf Def 8} also relies on counterfactuals, since it does not explicitly invoke counterfactual values of the candidate cause. Exploring this topic further lies beyond the scope of this paper.}

\begin{example}\label{counter} The equation for a binary variable $Y$ is such that $Y=1$ iff $N \neq 0$, and the range for $N$ is $\{0,1,2,3\}$. The equation for $N$ is as follows: $N=0$ if $A=0$, $N=1$ if $(A=1 \land X=1)$, $N=2$ if $(A=1 \land X=0 \land W=1)$, and $N=3$ if $(A=1 \land X=0 \land W=0)$. In a context where $A=W=X=1$, we get that $X=1$ causes $Y=1$ according to {\bf Def 2}. Yet there is no intervention such that $Y=1$ depends on $X=1$ under that intervention (and thus none of the other definitions would consider $X=1$ a cause of $Y=1$). In this case, both answers seem plausible. {\bf Def 2} reaches its verdict because of the asymmetry between 
$(A=1,X=1)$ and $(A=1,X=0)$: only the former is by itself causally sufficient for a network that results in $Y=1$, whereas the latter also needs the assistance of $W=1$ or $W=0$. 
\end{example}

Now we consider six examples which are simple variations on the same theme, because they all share the following equation for $Y$: $Y= (X \land D) \lor A$. Moreover, they all share a context such that $X=1$ and $A=1$. The only difference between them lies with the value of $D$ ($0$ or $1$) and with the relation between $A$ and $D$. (Concretely, there could be no relation, or it can be given by $A=D$, $A=\lnot D$, $D=A$, and $D=\lnot A$.) In all examples, all definitions agree that $A=1$ is a cause of $Y=1$. The disagreement arises over whether $X=1$ should be considered a cause as well. 

Intuitively, I would find it unacceptable to consider $X=1$ a cause whenever $D=0$, regardless of the relation between $A$ and $D$. The disjunct in which $X$ appears is false, and therefore it played no positive part whatsoever in causing $Y=1$. Perhaps others are more tolerant. But even if that is the case, one should expect one's verdicts to exhibit some consistency. As we will see, {\bf Def 2} and {\bf Original HP} are the only definitions which can meet this demand. 

The situation is simplest for {\bf Original HP}: it considers $X=1$ a cause of $Y=1$ no matter what. To see why, take as a witness $(D=1,A=0)$. Holding fixed that witness, $Y=1$ counterfactually depends on $X=1$. Since $\vec{Z}=\{X\}$, the former is equivalent to AC2 for {\bf Original HP}. So we gain consistency, but at the price of extreme tolerance. In fact, Halpern and Pearl use precisely this example to argue against {\bf Original HP} and in favor of {\bf Updated HP} (\citeyear[p. 35]{halpernpearl05a}):
\begin{example}\label{prisoner}
``Suppose that a prisoner dies either if $X$ loads $D$'s gun and $D$ shoots, or if $A$ loads and shoots his gun. Taking $Y$ to represent the prisoner's death and making the obvious assumptions about the meaning of the variables, ... [we can use the equation described above]. Suppose that $X$ loads $D$'s gun ($X=1$), $D$ does not shoot ($D=0$), but $A$ does load and shoot his gun ($A=1$), so that the prisoner dies. Clearly $A=1$ is a cause of $Y=1$. {\em We would not want to say that $X=1$ is a cause of $Y=1$, given that $D$ did not shoot (i.e., given that $D=0$).}'' [emphasis added]
\end{example}

If we agree with Halpern and Pearl here -- which I do -- then {\bf Original HP} can be discarded on the basis of this example (and on the basis of the many others we discussed previously, of course). I leave it to the reader to verify that none of the other definitions consider $X=1$ to be a cause here. 

However, the only definition that applies the intuition underlying this example to all cases in which $D=0$ is {\bf Def 2}. Moreover, it is the only remaining definition that offers a simple consistent answer in all cases: $X=1$ is a cause of $Y=1$ iff $D=1$. To see why this is the case, we go over the possible directly sufficient sets. (Since $X$ is only a parent of $Y$, we can invoke Proposition \ref{pro:parent} and use {\bf Def 3} instead of {\bf Def 2}.) Clearly $X=1$ is not directly sufficient for $Y=1$ by itself. It is also clear that we cannot add $A=1$ to the witness, because $A=1$ is directly sufficient for $Y=1$ all by itself. Therefore we are forced to choose $D$ as our witness. If $D=0$, this gives $(X=1,D=0)$, which is not directly sufficient for $Y=1$ and thus $X=1$ is not a cause. If $D=1$, we get $(X=1,D=1)$, which is directly sufficient for $Y=1$. Since the same does not hold for $(X=0, D=1)$, $X=1$ is a cause of $Y=1$.

The following examples show that {\bf Updated HP} and {\bf Modified HP} flip-flop between calling $X=1$ a cause or not even when holding fixed the value of $D$. Of course I cannot exclude the possibility that some consistent argumentation can be offered to explain the results of one of these definitions, but in its absence all of this speaks in favor of {\bf Def 2}. We start with the three possible ways in which it can arise that $D=1$.

\begin{example}\label{example1}
First consider the case where $D$ is determined by the context, and we have a context such that $D=1$. Here all four definitions agree that $X=1$ is a cause of $Y=1$.
\end{example}

\begin{example}\label{example2}
Second consider the case where the equation for $D$ is given by $D=A$ and thus again $D=1$ in the context under consideration. Here {\bf Updated HP} and {\bf Modified HP} flip their verdict, as they no longer consider $X=1$ a cause of $Y=1$.
\end{example}

\begin{example}\label{example3}
Third, we simply flip the relation between $A$ and $D$ so that $A=D$, and again $D=1$ in the context under consideration. Now {\bf Updated HP} and {\bf Modified HP} go back to considering $X=1$ a cause of $Y=1$.
\end{example}

Next we consider the two remaining possible cases where $D=0$ (Example \ref{prisoner} was the first such case).

\begin{example}\label{example4}
Consider the case where the equation for $D$ is $D= \lnot A$. As with Example \ref{prisoner}, we have that $D=0$, and yet {\bf Updated HP} changes its verdict, calling $X=1$ a cause of $Y=1$.
\end{example}

\begin{example}\label{example5}\footnote{The attentive reader will remember this example from the proof of Theorem \ref{thm:ident}.}
Lastly, consider the case where the equation for $D$ is $A=\lnot D$, and thus we again have that $D=0$. Now both {\bf Modified HP} and {\bf Updated HP} flip their verdicts as compared to Example \ref{prisoner}. To see why, it suffices to consider {\bf Modified HP}. The result for {\bf Updated HP} then follows from Theorem \ref{thm:relations}. $D=0$ by itself is not a cause of $Y=1$ because there is no choice of witness that makes $Y=1$ counterfactually depend on $D=0$. Since $Y=1$ does counterfactually depend on $(X=1,D=0)$, $X=1$ is part of a cause of $Y=1$. 
\end{example}

\section{Conclusion}\label{sec:con}

I have developed twelve definitions of actual causation that formalize the NESS intuition with which Pearl started, and have shown that the most recent of the HP definitions is among them. Although these definitions vary widely in terms of the verdicts they reach, they all resemble each other as being instantiations of the same general definition. Each definition is made up of two elements: a definition of causal sufficiency, and a definition of necessity. Other definitions can easily be developed by playing around with these elements. 

After studying various properties of these definitions and the relations between them, I moved on to the process of selecting the definition that does best in practice. In the majority of the many examples that we have considered, {\bf Def 2} agrees with {\bf Modified HP}. However, in Section \ref{sec:modhp} we came across two examples for which {\bf Def 2} disagreed with {\bf Modified HP} and where {\bf Modified HP} gave the wrong verdict. Moreover, contrary to {\bf Modified HP}, {\bf Def 2} manages to give consistent (and intuitive) answers to the group of cases considered in the previous section. Therefore I conclude by suggesting that we should adopt {\bf Def 2} as a definition of actual causation. This definition is made up of strong sufficiency and contrastive necessity. It states that $\vec{X}=\vec{x}$ causes $Y=y$ iff $\vec{X}=\vec{x}$ is a Minimal Contrastively Necessary Subset of a Strongly Sufficient Set for $Y=y$, or $\text{MCNS}^4$.

\proofs{
\setcounter{section}{0}
\renewcommand\thesection{\Alph{section}}

\section{Appendix}

\section*{Causal Sufficiency}

\opro{pro:network} $\vec{X}=\vec{x}$ is strongly sufficient for $\vec{Y}=\vec{y}$ in $M$ along a network $\vec{N}$ iff $\vec{X}=\vec{x}$ is strongly sufficient for $\vec{Y}=\vec{y}$ in $M$.
\eopro

\prf First assume $\vec{X}=\vec{x}$ is strongly sufficient for $\vec{Y}=\vec{y}$ in $M$ and $\vec{N}$ can be used to show this. Then the result follows immediately from the observation that $\vec{X}=\vec{x}$ is directly sufficient for $\vec{N}=\vec{n}$ and either $\vec{N}=\vec{n}$ is directly sufficient for $\vec{Y}=\vec{y}$ or $\vec{N}=\vec{Y}$ and  $\vec{n}=\vec{y}$.

Second assume $\vec{X}=\vec{x}$ is strongly sufficient for $Y=y$ in $M$ along a network $\vec{N}$. Define $\vec{A}=\V \minus (\vec{X} \cup \vec{N})$. We need to show that for all $\vec{a} \in \R(\vec{A})$ and all $\vec{u} \in \R(\U)$ we have that $(M,\vec{u})\sat [\vec{X} \gets \vec{x},\vec{A} \gets \vec{a}] \vec{N}=\vec{n}$. 

We know that $\vec{X}=\vec{x}$ is directly sufficient for $\vec{N_1}=\vec{n_1}$. Define $\vec{C_1}=\V - (\vec{X} \cup \vec{N_1})$ and $\vec{D_1}=\vec{N} \minus \vec{N_1}$. Note that $\vec{C_1}=\vec{A} \cup \vec{D_1}$. We have that for all $\vec{c_1} \in \R(\vec{C_1})$ and all $\vec{u} \in \R(\U)$, $(M,\vec{u})\sat [\vec{X} \gets \vec{x},\vec{C_1} \gets \vec{c_1}] \vec{N_1}=\vec{n_1}$. In particular, we have that for all $\vec{a} \in \R(\vec{A})$ and all $\vec{u} \in \R(\U)$, $(M,\vec{u})\sat [\vec{X} \gets \vec{x}, \vec{A} \gets \vec{a}] \vec{N_1}=\vec{n_1}$.

Define $\vec{C_2}=\V - (\vec{N_1} \cup \vec{N_2})$ and $\vec{D_2}=\vec{N} \minus (\vec{N_1} \cup \vec{N_2})$. Note that $\vec{C_2}=\vec{A} \cup \vec{D_2} \cup \vec{X}$. We have that for all $\vec{c_2} \in \R(\vec{C_2})$ and all $\vec{u} \in \R(\U)$, $(M,\vec{u})\sat [\vec{N_1} \gets \vec{n_1},\vec{C_2} \gets \vec{c_2}] \vec{N_2}=\vec{n_2}$. In particular, we have that for all $\vec{a} \in \R(\vec{A})$ and all $\vec{u} \in \R(\U)$, $(M,\vec{u})\sat [\vec{X} \gets \vec{x}, \vec{N_1} \gets \vec{n_1},\vec{A} \gets \vec{a}] \vec{N_2}=\vec{n_2}$. Combined with the conclusion from the previous paragraph, it follows that for all $\vec{a} \in \R(\vec{A})$ and all $\vec{u} \in \R(\U)$, $(M,\vec{u})\sat [\vec{X} \gets \vec{x}, \vec{A} \gets \vec{a}] \vec{N_1}=\vec{n_1} \land \vec{N_2}=\vec{n_2}$.

Defining $\vec{N_{k+1}}=\vec{Y}$, we can generalize this reasoning for all consecutive $i \in \{3,\ldots,k+1\}$ to get the desired outcome.
\eprf

\commentout{
\opro{pro:suf}
The three variants of sufficiency are equivalent to Definition \ref{def:gensuf} when making the following choices for $\vec{N}$ and $\vec{C}$:
\begin{description}
\item[{\rm Weak Sufficiency.}] Choose both $\vec{C}$ and $\vec{N}$ to be minimal, i.e., $\vec{C}=\vec{N}=\emptyset$.
\item[{\rm Strong Sufficiency.}] Choose $\vec{N}$ to be maximal given $\vec{C}$, i.e., $\vec{N}=\V \minus (\vec{X} \cup \vec{Y} \cup \vec{C})$.
\item[{\rm Direct Sufficiency.}] Choose $\vec{C}$ to be maximal, i.e., $\vec{C} = \V \minus (\vec{X} \cup \vec{Y})$ and thus $\vec{N}=\emptyset$.
\end{description}
\eopro

\prf Proposition \ref{pro:suf} is trivial for both Weak and Direct Sufficiency. For Strong Sufficiency, note that the first statement in Definition \ref{def:gensuf} simply expresses that $(\vec{X}=\vec{x},\vec{N}=\vec{n})$ is directly sufficient for $\vec{Y}=\vec{y}$, and the second statement expresses that $\vec{X}=\vec{x}$ is directly sufficient for $\vec{N}=\vec{n}$. Taken together, this means precisely that $\vec{X}=\vec{x}$ is strongly sufficient for $\vec{Y}=\vec{y}$ along $\vec{N}$.
\eprf
}

\commentout{
\opro{pro:sim}
$\vec{X}=\vec{x}$ is strongly sufficient for $Y=y$ iff there exists some $\vec{N} \subseteq (\V \minus (\vec{X} \cup \{Y\}))$ and a setting $\vec{n} \in \R(\vec{N})$ such that for all $\vec{c} \in \R(\V \minus (\vec{N} \cup \vec{X} \cup \{Y\}))$ and for all $\vec{u} \in \R(\U)$ we have that $(M,\vec{u})\sat [\vec{X} \gets \vec{x},\vec{C} \gets \vec{c}, \vec{N} \gets \vec{n}]Y=y$ and $(M,\vec{u})\sat [\vec{X} \gets \vec{x},\vec{C} \gets \vec{c}] \vec{N}=\vec{n}$.
\eopro

\prf Given strong recursivity, it is impossible that $Y$ depends on any of its descendants. Therefore we can always choose $\vec{N}$ so that it does not contain any descendants of $Y$ in any context. In turn, it is impossible that any member of $\vec{N}$ depends on $Y$ in any context. From this the result follows.
\eprf
}

\section*{Defining Causation using Sufficiency}

\commentout{
\othm{thm:ident} The following are all equivalences among the eight definitions and the three HP definitions:
\begin{itemize}
	\item  {\bf Modified HP} iff {\bf Def 1}
	\item  {\bf Def 4} iff {\bf Def 8}
\end{itemize}
\eothm 

\prf First we consider the equivalences that do hold. 

We start with the first equivalence: {\bf Modified HP} iff {\bf Def 1}. This is simply a matter of explicitly writing out the definitions, starting with actual weak sufficiency: $\vec{X}=\vec{x}$ is actually weakly sufficient for $Y=y$ in $(M,\vec{u})$ iff  $(M,\vec{u})\sat [\vec{X} \gets \vec{x}] Y=y$. Next we note that the following condition is trivially satisfied for any $\vec{W} \subseteq \V$: $(M,\vec{u})\sat [\vec{X} \gets \vec{x},\vec{W} \gets \vec{w}^*] Y=y$. 

Combining both claims, we can rewrite {\bf Modified HP} as follows, which gives the desired result:
\begin{description}
\item[{\rm AC2(a).}] There is a set $\vec{W} \subseteq (\V \minus (\vec{X} \cup \{Y\}))$ and a setting $\vec{x}'$ of the variables in $\vec{X}$ such that 
$(\vec{X}=\vec{x}', \vec{W}=\vec{w}^*)$ is not actually weakly sufficient for $Y=y$ in $(M,\vec{u})$.
\item[{\rm AC2(b).}] $(\vec{X}=\vec{x}, \vec{W}=\vec{w}^*)$ is actually weakly sufficient for $Y=y$ in $(M,\vec{u})$.
\end{description}

Now we prove the second equivalence: {\bf Def 4} iff {\bf Def 8}. We need to show that the following two statements are equivalent:
\begin{itemize}
	\item $\vec{W}=\vec{w}^*$ is not directly sufficient for $Y=y$.
	\item There exists values $\vec{x}'$ of $\vec{X}$ such that $(\vec{X}=\vec{x}',\vec{W}=\vec{w}^*)$ is not directly sufficient for $Y=y$.
\end{itemize}

Filling in Definition \ref{def:dirsuf}, the result follows immediately:
\begin{itemize}
	\item There exists a $\vec{z} \in \R(\V - (\vec{W} \cup \vec{X} \cup \{Y\}))$, a $\vec{x}' \in \R(\vec{X})$, and a $\vec{u}' \in \R(\U)$ so that $(M,\vec{u}')\sat [\vec{W} \gets \vec{w}^*, \vec{X} \gets \vec{x}',\vec{C} \gets \vec{c}] Y \neq y$.
	\item There exists values $\vec{x}'$ of $\vec{X}$, a $\vec{z} \in \R(\V - (\vec{W} \cup \vec{X} \cup \{Y\}))$ and a $\vec{u}' \in \R(\U)$ so that $(M,\vec{u}')\sat [\vec{W} \gets \vec{w}^*, \vec{X} \gets \vec{x}',\vec{C} \gets \vec{c}] Y \neq y$. 
\end{itemize} 

Second, we go over some examples to show that none of the other equivalences hold. (Obviously, from now on we may ignore {\bf Def 1}, {\bf Def 7*}, and {\bf Def 8}.)  

\begin{example}\label{ex1} Equations: $Y=(X \land A) \lor D$, $D=A$. Context: $A=1$. Then $X=1$ is a cause of $Y=1$ according to: 
	\begin{itemize}
		\item {\bf Modified HP}: $Y=1$ counterfactually depends on $(X=1,D=1)$, and not on either $X=1$ or $D=1$. So $X=1$ is part of a cause.
		\item {\bf Updated HP} and {\bf Original}: take $(A=1,D=0)$ as a witness.
		\item {\bf Def 4}: again take $(A=1,D=0)$ as a witness.
		\item {\bf Def 3}: follows from the previous item and Theorem \ref{thm:relations}.
		\item {\bf Def 7}: follows from the previous item and Theorem \ref{thm:relations}.
	\end{itemize}
		
$X=1$ is not a cause of $Y=1$ according to:
	\begin{itemize}
		\item {\bf Def 10*}: $X=1$ by itself does not weakly suffice for $Y=1$ (just look at a context in which $A=0$), so we need to add $A$ or $D$ to the witness. But both $A=1$ and $D=1$ each weakly suffice for $Y=1$.
		\item {\bf Def 2}: $(X=0,A=1)$ and $(X=0,D=1)$ also weakly suffice for $Y=1$.
	\end{itemize}
\end{example}

So we know that {\bf Def 2} and {\bf Def 10*} are not equivalent to any of the other definitions. We give an example to show that {\bf Def 2} and {\bf Def 10*} are not equivalent to each other either. 

\begin{example}\label{ex2} Equations: $Y=X \land A$, $X=A$. Context: $A=1$. Since $X=1$ is not weakly sufficient for $Y=1$, we need to include $A=1$ in the witness. Indeed, $(X=1,A=1)$ is weakly sufficient for $Y=1$. However, so is $A=1$, and therefore $X=1$ does not cause $Y=1$ according to {\bf Def 10*}. Yet $(X=0,A=1)$ is not weakly sufficient for $Y=1$, and therefore $X=1$ causes $Y=1$ according to {\bf Def 2}.
\end{example}

This leaves us with the HP definitions, {\bf Def 3}, {\bf Def 4}, and {\bf Def 7}. The next example shows that the former are not equivalent to the latter. 
	
\begin{example}\label{ex3} Equations: $Y=(X \land \lnot A) \lor D$, $D=A$. Context: $A=1$. Then $X=1$ is a cause of $Y=1$ according to: 
	\begin{itemize}
		\item {\bf Modified HP}: $Y=1$ counterfactually depends on $(X=1,A=1)$, and not on either $X=1$ or $A=1$. So $X=1$ is part of a cause.
		\item {\bf Updated HP} and {\bf Original}: take $A=0$ as a witness.
	\end{itemize}
		
$X=1$ is not a cause of $Y=1$ according to:
	\begin{itemize}
		\item {\bf Def 4}: $X=1$ by itself does not directly suffice for $Y=1$ (just look at $[A \gets 1, D \gets 0]$), so we need to add $A$ or $D$ to the witness. Since the actual value of $A$ is $1$, it is of no use, which leaves us with $D$. But $D=1$ directly suffices for $Y=1$ by itself, and thus so does $(X=0,D=1)$.
		\item {\bf Def 3}: follows from the previous item and Proposition \ref{pro:parent}.
		\item {\bf Def 7}: follows from the previous item and Proposition \ref{pro:parent}.
	\end{itemize}
\end{example}

That none of the HP definitions are equivalent is of course a well-established fact, and also follows from the examples we consider in Section \ref{sec:237}. Therefore we are left with showing that {\bf Def 3}, {\bf Def 4}, and {\bf Def 7} are not equivalent. That {\bf Def 4} differs from the other two is a direct consequence of some of our later results, but a simple example illustrates this as well. 

\begin{example}\label{ex4} Equations: $Y=A$, $A=X$. Context: $A=1$. Then it is easy to see that $X=1$ causes $Y=1$ according to all definitions here considered, except for {\bf Def 4}.
\end{example}

Lastly, I refer the reader to Example \ref{switch} in Sections \ref{sec:237} for an example that shows {\bf Def 3} and {\bf Def 7} are not equivalent.
\commentout{
Lastly, here is an example that shows {\bf Def 3} and {\bf Def 7} are not equivalent.
\begin{example}\label{ex37} 
The equation for a binary variable $Y$ is such that $Y=1$ iff $N \neq 0$, and the range for $N$ is $\{0,1,2,3\}$. We also have a binary variable $A$ and a variable $X$ with range $\{0,1,2\}$. The equation for $N$ is as follows: $N=0$ if $A=0$, $N=1$ if $(A=1 \land X=1)$, $N=2$ if $(A=1 \land X=0)$, and $N=3$ if $(A=1 \land X=2)$. Consider a context where $A=X=1$. Then $(A=1,X=1)$ is strongly sufficient for $Y=1$ along $\{N\}$, whereas $A=1$ is not strongly sufficient for $Y=1$ along any subnetwork. Therefore $X=1$ causes $Y=1$ according to  {\bf Def 7}. Yet both $(A=1,X=0)$ and $(A=1,X=2)$ are also strongly sufficient for $Y=1$ along $\{N\}$, and thus {\bf Def 3} does not consider $X=1$ a cause of $Y=1$.
\end{example}
}
\eprf
}

\othm{thm:ident} The following are all equivalences among the twelve definitions and the three HP definitions:
\begin{itemize}
	\item  {\bf Modified HP} iff {\bf Def 1}
	\item  {\bf Def 2} iff {\bf Def 5}
	\item  {\bf Def 8} iff {\bf Def 11}
	\item  {\bf Def 3} iff {\bf Def 6} iff {\bf Def 9} iff {\bf Def 12}
\end{itemize}
\eothm 

\prf First we consider the equivalences that do hold. 

We start with the first equivalence: {\bf Modified HP} iff {\bf Def 1}. This is simply a matter of explicitly writing out the definitions, starting with actual weak sufficiency: $\vec{X}=\vec{x}$ is actually weakly sufficient for $Y=y$ in $(M,\vec{u})$ iff  $(M,\vec{u})\sat [\vec{X} \gets \vec{x}] Y=y$. Next we note that the following condition is trivially satisfied for any $\vec{W} \subseteq \V$: $(M,\vec{u})\sat [\vec{X} \gets \vec{x},\vec{W} \gets \vec{w}^*] Y=y$. 

Combining both claims, we can rewrite {\bf Modified HP} as follows, which gives the desired result:
\begin{description}
\item[{\rm AC2(a).}] There is a set $\vec{W} \subseteq (\V \minus (\vec{X} \cup \{Y\}))$ and a setting $\vec{x}'$ of the variables in $\vec{X}$ such that 
$(\vec{X}=\vec{x}', \vec{W}=\vec{w}^*)$ is not actually weakly sufficient for $Y=y$ in $(M,\vec{u})$.
\item[{\rm AC2(b).}] $(\vec{X}=\vec{x}, \vec{W}=\vec{w}^*)$ is actually weakly sufficient for $Y=y$ in $(M,\vec{u})$.
\end{description}

Next we consider all of the following equivalences: {\bf Def 2} iff {\bf Def 5}, {\bf Def 8} iff {\bf Def 11}, {\bf Def 3} iff {\bf Def 6}, {\bf Def 9} iff {\bf Def 12}. The reason we can group these together, is because we can prove all of them by invoking the following observation and two subsequent lemmas.

\begin{observation}\label{obs:obs1} Recall our restriction on causal models that exogenous variables only appear in equations of the form $V=U$. Say $\vec{R} \subseteq \V$ are all variables which have such an equation, and call these the {\em root} variables. It is clear that if we intervene on all of the root variables, they take over the role of the exogenous variables. Concretely, given strong recursivity, for any setting $\vec{r} \in \R(\vec{R})$ there exists a unique setting $\vec{v} \in \R(\V)$ so that for all contexts $\vec{u} \in \R(\U)$ we have that $(M,\vec{u}) \sat [\vec{R} \gets \vec{r}]\V=\vec{v}$.
\end{observation}

\begin{lemma}\label{lem:actual} Given a setting $\vec{X}=\vec{x}$, a setting $\vec{N}=\vec{n}$ that includes $Y=y$ and such that $\vec{N} \cap \vec{R}=\emptyset$, a context $\vec{u}$, the following holds:\footnote{$\vec{R}$ is defined in Observation \ref{obs:obs1}.}
	\begin{itemize}
		\item $\vec{X}=\vec{x}$ is actually directly sufficient for $Y=y$ in $(M,\vec{u})$ iff $\vec{X}=\vec{x}$ is directly sufficient for $Y=y$ in $M$;
		\item $\vec{X}=\vec{x}$ is actually strongly sufficient for $Y=y$ in $(M,\vec{u})$ along $\vec{N}=\vec{n}$ iff $\vec{X}=\vec{x}$ is strongly sufficient for $Y=y$ in $M$ along $\vec{N}=\vec{n}$.
	\end{itemize}
\end{lemma}

\prf Filling in the definitions of direct and actually direct sufficiency, the first equivalence reduces to the following: for all $\vec{c} \in \R(\V \minus (\vec{X} \cup \{Y\}))$, it holds that $(M,\vec{u}) \sat [\vec{X} \gets \vec{x}, \vec{C} \gets \vec{c}]Y=y$ iff for all $\vec{u}'' \in \R(\U)$, $(M,\vec{u}'') \sat [\vec{X} \gets \vec{x}, \vec{C} \gets \vec{c}]Y=y$.

Because of Observation \ref{obs:obs1}, we have that for any setting $\vec{v} \in \V$ and any setting $\vec{r} \in \R(\vec{R})$, it holds that $(M,\vec{u}) \sat [\vec{R} \gets \vec{r}]\V=\vec{v}$ iff for all contexts $\vec{u}'' \in \R(\U)$, $(M,\vec{u}'') \sat [\vec{R} \gets \vec{r}]\V=\vec{v}$. Combining this with the fact that $\vec{R} \subseteq (\vec{C} \cup \vec{X})$ gives the desired result.

The second equivalence can be reformulated as follows: $\vec{X}=\vec{x}$ is actually directly sufficient for $\vec{N}=\vec{n}$ in $(M,\vec{u})$ iff $\vec{X}=\vec{x}$ is directly sufficient for $\vec{N}=\vec{n}$ in $M$. In turn, this reduces to: for all $\vec{c} \in \R(\V \minus (\vec{X} \cup \vec{N}))$, it holds that $(M,\vec{u}) \sat [\vec{X} \gets \vec{x}, \vec{C} \gets \vec{c}]\vec{N}=\vec{n}$ iff for all $\vec{u}'' \in \R(\U)$, $(M,\vec{u}'') \sat [\vec{X} \gets \vec{x}, \vec{C} \gets \vec{c}]\vec{N}=\vec{n}$.

Given that $\vec{N} \cap \vec{R}=\emptyset$, we still have that $\vec{R} \subseteq (\vec{C} \cup \vec{X})$, and therefore we can apply the same reasoning as before.
\eprf

\begin{lemma}\label{lem:noroot} For all twelve instances of the {\bf General Definition of Causation} we can restrict ourselves to sets $\vec{N}$ so that $(\vec{N} \minus \{Y\}) \cap \vec{R}=\emptyset$.
\end{lemma} 

\prf Let $\vec{A}$ denote $(\vec{N} \minus \{Y\}) \cap \vec{R}$. For all definitions using either variants of direct or weak sufficiency the result follows immediately from the fact that $\vec{N} \minus \{Y\}=\emptyset$.

First consider the case where we use non-actual strong sufficiency ({\bf Def 5} or {\bf Def 11}). In that case, AC2(b) can never be satisfied unless $\vec{A}=\emptyset$. To see why, note that in all contexts $\vec{u}'' \in \R(\U)$, it has to hold that $(M,\vec{u}'')\sat [\vec{X} \gets \vec{x}, \vec{W} \gets \vec{w}^*] \vec{A}=\vec{a}$. Since $\vec{A} \cap (\vec{X} \cup \vec{W})$ and the equation for each element $A_i \in \vec{A}$ is of the form $A_i = U$ for some exogenous variable $U$, this is impossible. (Strictly speaking it is possible, namely if the range of $U$ consists only of the single value $a_i^*$. Although I did not make this explicit in Section \ref{sec:equations}, it is standard to assume that all variables have a range that contains at least two elements.) 

Second consider the case where we use actual strong sufficiency and contrastive necessity ({\bf Def 2}). (The case of {\bf Def 8} is entirely analogous.) Say we are considering a candidate cause $\vec{X}=\vec{x}$, a candidate witness $\vec{W}=\vec{w}^*$, contrast values $\vec{x}'$, and a setting $\vec{N}=\vec{n}$ that includes $Y=y$. Given AC1, we can safely assume that $\vec{n}=\vec{n}^*$.

I claim that the following holds, from which the result follows: $\vec{X}=\vec{x}$ satisfies AC2 using contrast values $\vec{x}'$, witness $\vec{W}=\vec{w}^*$, and network $\vec{N}$ iff $\vec{X}=\vec{x}$ satisfies AC2 using contrast values $\vec{x}'$, witness $(\vec{W}=\vec{w}^*,\vec{A}=\vec{a}^*)$, and network $\vec{N} \minus \vec{A}$.

Because $\vec{A} \subseteq \vec{R}$, we have that for any set $\vec{B} \subseteq (\V \minus \vec{A})$, and any setting $\vec{b} \in \R(\vec{B})$, $(M,\vec{u}) \sat [\vec{B} \gets \vec{b}]\vec{A}=\vec{a}^*$. Moreover, since $(M,\vec{u}) \sat \vec{A}=\vec{a}^*$, for each setting $\vec{v} \in (\V \minus \vec{A})$ we also have that $(M,\vec{u}) \sat [\vec{B} \gets \vec{b}](\V \minus \vec{A})=\vec{v}$ iff $(M,\vec{u}) \sat [\vec{B} \gets \vec{b},\vec{A} \gets \vec{a}^*](\V \minus \vec{A})=\vec{v}$.

Using these observations and the fact that $\vec{A} \subseteq \vec{N}$, we get that the following two conditions are equivalent, for which the result follows as far as AC2(b) is concerned:
\begin{description}
\item[{\rm AC2(b).}] For all $\vec{c} \in \R(\V - (\vec{X} \cup \vec{W} \cup \vec{N}))$ we have that $(M,\vec{u})\sat [\vec{X} \gets \vec{x}, \vec{W} \gets \vec{w}^*, \vec{C} \gets \vec{c}] \vec{N}=\vec{n}^*$. 
\item[{\rm AC2(b).}] For all $\vec{c} \in \R(\V - (\vec{X} \cup \vec{W} \cup \vec{N}))$ we have that $(M,\vec{u})\sat [\vec{X} \gets \vec{x}, \vec{W} \gets \vec{w}^*, \vec{A} \gets \vec{a}^*, \vec{C} \gets \vec{c}] (\vec{N} \minus \vec{A})=\vec{n_2}^*$ (where $\vec{n_2}$ is the restriction of $\vec{n}^*$ to $(\vec{N} \minus \vec{A})$).
\end{description}

Now we focus on AC2($\text{a}^{\text{c}}$). 

Let us first assume AC2($\text{a}^{\text{c}}$) holds for $\vec{X}=\vec{x}$, contrast values $\vec{x}'$, witness $(\vec{W}=\vec{w}^*,\vec{A}=\vec{a}^*)$, and network $\vec{N} \minus \vec{A}$. We need to show that it holds for $\vec{X}=\vec{x}$, contrast values $\vec{x}'$, witness $(\vec{W}=\vec{w}^*)$, and network $\vec{N}$.

Consider some $\vec{S} \subseteq \vec{N}$ with $Y \in \vec{S}$. We need to find a $\vec{t} \in \R(\V \minus (\vec{X} \cup \vec{W} \cup \vec{S}))$ so that 
$(M,\vec{u})\sat [\vec{X} \gets \vec{x}',\vec{W} \gets \vec{w}^*,\vec{T} \gets \vec{t}] \vec{S} \neq \vec{s}^*$. Define $\vec{S_1}=\vec{S} \minus \vec{A}$, $\vec{S_2}=\vec{S} \cap \vec{A}$, and $\vec{A_1}=\vec{A} \minus \vec{S}$.

Since $\vec{S_1} \subseteq (\vec{N} \minus \vec{A})$ with $Y \in \vec{S_1}$, we know that there exists some $\vec{t_1} \in \R(\V \minus (\vec{X} \cup \vec{W} \cup \vec{A} \cup \vec{S_1})$ so that 
$(M,\vec{u})\sat [\vec{X} \gets \vec{x}',\vec{W} \gets \vec{w}^*,\vec{A} \gets \vec{a}^*, \vec{T} \gets \vec{t_1}] \vec{S_1} \neq \vec{s_1}^*$. Since $\vec{S_1} \subseteq \vec{S}$, it also holds that $(M,\vec{u})\sat [\vec{X} \gets \vec{x}',\vec{W} \gets \vec{w}^*,\vec{A} \gets \vec{a}^*, \vec{T} \gets \vec{t_1}] \vec{S} \neq \vec{s}^*$. Also, given our observations about $\vec{A}$, it also follows that  $(M,\vec{u})\sat [\vec{X} \gets \vec{x}',\vec{W} \gets \vec{w}^*, \vec{A_1} \gets \vec{a_1}, \vec{T} \gets \vec{t_1}] \vec{S} \neq \vec{s}^*$. Lastly, note that $[\V \minus (\vec{X} \cup \vec{W} \cup \vec{A} \cup \vec{S_1})] \cup \vec{A_1}=\V \minus (\vec{X} \cup \vec{W} \cup \vec{S})$. Therefore we can choose $\vec{t}=(\vec{a_1},\vec{t_1})$. 

Next we consider the other direction: assume AC2($\text{a}^{\text{c}}$) holds for $\vec{X}=\vec{x}$, contrast values $\vec{x}'$, witness $\vec{W}=\vec{w}^*$, and network $\vec{N}$. We need to show that it holds for $\vec{X}=\vec{x}$, contrast values $\vec{x}'$, witness $(\vec{W}=\vec{w}^*,\vec{A}=\vec{a}^*)$, and network $\vec{N} \minus \vec{A}$.

Consider some $\vec{S} \subseteq (\vec{N} \minus \vec{A})$ with $Y \in \vec{S}$. We need to find a $\vec{t} \in \R(\V \minus (\vec{X} \cup \vec{W} \cup \vec{A} \cup \vec{S})$ so that 
$(M,\vec{u})\sat [\vec{X} \gets \vec{x}',\vec{W} \gets \vec{w}^*,\vec{A} \gets \vec{a}^*,\vec{T} \gets \vec{t}] \vec{S} \neq \vec{s}^*$. 

Note that $(\vec{S} \cup \vec{A}) \subseteq \vec{N}$, and also $Y \in (\vec{S} \cup \vec{A})$. Therefore there exists some $\vec{t_2} \in \R(\V \minus (\vec{X} \cup \vec{W} \cup \vec{A} \cup \vec{S})$ so that 
$(M,\vec{u})\sat [\vec{X} \gets \vec{x}',\vec{W} \gets \vec{w}^*,\vec{A} \gets \vec{a}^*, \vec{T} \gets \vec{t_2}] (\vec{S} \neq \vec{s}^* \lor \vec{A} \neq \vec{a}^*)$. It follows that $(M,\vec{u})\sat [\vec{X} \gets \vec{x}',\vec{W} \gets \vec{w}^*,\vec{A} \gets \vec{a}^*, \vec{T} \gets \vec{t_2}] \vec{S} \neq \vec{s}^*$. Choosing $\vec{t}=\vec{t_2}$ gives the desired result.
\eprf

Because of the above lemmas, all that remains is to show that the above equivalences hold also when $Y \in \vec{R}$. This is accomplished by showing that settings of such variables do not have any cause, regardless of the definition one uses. 

AC2(a) requires us to look at all subsets of $\vec{N}=\vec{n}$ that include $Y=y$, and verify that the candidate cause and witness $(\vec{X}=\vec{x}',\vec{W}=\vec{w}^*)$ (or candidate witness $\vec{W}=\vec{w}^*$ in case we use AC2($\text{a}^{\text{m}}$)) is not sufficient for that subset. One such subset is the one containing just $Y=y$. By AC1, we have that $(M,\vec{u}) \sat Y=y$. Since $Y \in \vec{R}$, there is no intervention on the other endogenous variables so that $Y \neq y$ under that intervention in $\vec{u}$. Therefore any definition of causation using a version of {\em actual} sufficiency (i.e., {\bf Def 2}, {\bf Def 3}, {\bf Def 8}, and {\bf Def 9}) considers all sets that do not include $Y$ to be sufficient for $Y=y$ in $(M,\vec{u})$. In particular, they consider $(\vec{X}=\vec{x}',\vec{W}=\vec{w}^*)$ to be sufficient for $Y=y$ in $(M,\vec{u})$, and thus fail to meet condition AC2(a).

For the definitions using {\em non-actual} variants of sufficiency ({\bf Def 5}, {\bf Def 6}, {\bf Def 11}, and {\bf Def 12}), it is condition AC2(b) that can never be satisfied. Analogous to what we saw in the proof of Lemma \ref{lem:noroot}, this follows from the fact that whatever version of sufficiency we use, $Y=y$ has to hold in all contexts, which is impossible given that $Y \not \in (\vec{X} \cup \vec{W})$. From this the result follows.

Now we prove the only remaining equivalence: {\bf Def 6} iff {\bf Def 12}. (Given the previous equivalences, other choices are possible too.) We need to show that the following two statements are equivalent:
\begin{itemize}
	\item $\vec{W}=\vec{w}^*$ is not directly sufficient for $Y=y$.
	\item There exists values $\vec{x}'$ of $\vec{X}$ such that $(\vec{X}=\vec{x}',\vec{W}=\vec{w}^*)$ is not directly sufficient for $Y=y$.
\end{itemize}

Filling in Definition \ref{def:dirsuf}, the result follows immediately:
\begin{itemize}
	\item There exists a $\vec{z} \in \R(\V - (\vec{W} \cup \vec{X} \cup \{Y\}))$, a $\vec{x}' \in \R(\vec{X})$, and a $\vec{u}' \in \R(\U)$ so that $(M,\vec{u}')\sat [\vec{W} \gets \vec{w}^*, \vec{X} \gets \vec{x}',\vec{C} \gets \vec{c}] Y \neq y$.
	\item There exists values $\vec{x}'$ of $\vec{X}$, a $\vec{z} \in \R(\V - (\vec{W} \cup \vec{X} \cup \{Y\}))$ and a $\vec{u}' \in \R(\U)$ so that $(M,\vec{u}')\sat [\vec{W} \gets \vec{w}^*, \vec{X} \gets \vec{x}',\vec{C} \gets \vec{c}] Y \neq y$. 
\end{itemize} 

Second, we go over some examples to show that none of the other equivalences hold. (Obviously, from now on we may ignore {\bf Def 1}, {\bf Def 5}, {\bf Def 6}, {\bf Def 7},  {\bf Def 9}, {\bf Def 11}, and {\bf Def 12}.)

\begin{example}\label{ex1} Equations: $Y=(X \land A) \lor D$, $D=A$. Context: $A=1$. Then $X=1$ is a cause of $Y=1$ according to: 
	\begin{itemize}
		\item {\bf Modified HP}: We can always consider choosing $\vec{W}=\emptyset$, in which case we simply get counterfactual dependence: $(M,\vec{u}) \sat \vec{X}=\vec{x} \land Y=1$ and $(M,\vec{u}) \sat [\vec{X} \gets \vec{x}']Y \neq y$. Doing so in this example, we see that $Y=1$ counterfactually depends on $(X=1,D=1)$. There is clearly also no witness $\vec{W}=\vec{w}^*$ to show that $X=1$ or $D=1$ are causes by themselves, so $X=1$ is part of a cause.
		\item {\bf Updated HP} and {\bf Original HP}: taking $(A=1,D=0)$ as a witness meets the conditions.
		\item {\bf Def 3}: again take $(A=1,D=0)$ as a witness.
		\item {\bf Def 2}: follows from the previous item and Theorem \ref{thm:relations}.
		\item {\bf Def 8}: follows from the previous item and Theorem \ref{thm:relations}.
	\end{itemize}
	
$X=1$ is not a cause of $Y=1$ according to:
	\begin{itemize}
		\item {\bf Def 10}: $X=1$ by itself does not weakly suffice for $Y=1$ (just look at a context in which $A=0$), so we need to add $A$ or $D$ to the witness. But both $A=1$ and $D=1$ each weakly suffice for $Y=1$.
		\item {\bf Def 4}: $(X=0,A=1)$ and $(X=0,D=1)$ also weakly suffice for $Y=1$.
	\end{itemize}
\end{example}

So we know that {\bf Def 4} and {\bf Def 10} are not equivalent to any of the other definitions. We give an example to show that {\bf Def 4} and {\bf Def 10} are not equivalent to each other either. 

\begin{example}\label{ex2} Equations: $Y=X \land A$, $X=A$. Context: $A=1$. Since $X=1$ is not weakly sufficient for $Y=1$, we need to include $A=1$ in the witness. Indeed, $(X=1,A=1)$ is weakly sufficient for $Y=1$. However, so is $A=1$, and therefore $X=1$ does not cause $Y=1$ according to {\bf Def 10}. Yet $(X=0,A=1)$ is not weakly sufficient for $Y=1$, and therefore $X=1$ causes $Y=1$ according to {\bf Def 4}.
\end{example}

This leaves us with the HP definitions, {\bf Def 2}, {\bf Def 3}, and {\bf Def 8}. The next example shows that the former are not equivalent to the latter. 
	
\begin{example}\label{ex3} Equations: $Y=(X \land \lnot A) \lor D$, $D=A$. Context: $A=1$. Then $X=1$ is a cause of $Y=1$ according to: 
	\begin{itemize}
		\item {\bf Modified HP}: $Y=1$ counterfactually depends on $(X=1,A=1)$, and not on either $X=1$ or $A=1$. So $X=1$ is part of a cause.
		\item {\bf Updated HP} and {\bf Original}: take $A=0$ as a witness.
	\end{itemize}
		
$X=1$ is not a cause of $Y=1$ according to:
	\begin{itemize}
		\item {\bf Def 3}: $X=1$ by itself does not directly suffice for $Y=1$ (just look at $[A \gets 1, D \gets 0]$), so we need to add $A$ or $D$ to the witness. Since the actual value of $A$ is $1$, it is of no use, which leaves us with $D$. But $D=1$ directly suffices for $Y=1$ by itself, and thus so does $(X=0,D=1)$.
		\item {\bf Def 2}: follows from the previous item and Proposition \ref{pro:parent}.
		\item {\bf Def 8}: follows from the previous item and Proposition \ref{pro:parent}.
	\end{itemize}
\end{example}

That none of the HP definitions are equivalent is of course a well-established fact, and also follows from the examples we consider in Section \ref{sec:237}. Therefore we are left with showing that {\bf Def 2}, {\bf Def 3}, and {\bf Def 8} are not equivalent. That {\bf Def 3} differs from the other two is a direct consequence of some of our later results, but a simple example illustrates this as well. 

\begin{example}\label{ex4} Equations: $Y=A$, $A=X$. Context: $A=1$. Then it is easy to see that $X=1$ causes $Y=1$ according to all definitions here considered, except for {\bf Def 3}.
\end{example}

Lastly, I refer the reader to Example \ref{switch} in Sections \ref{sec:237} for an example that shows {\bf Def 2} and {\bf Def 8} are not equivalent.
\eprf

\opro{pro:single} If $\vec{X}=\vec{x}$ causes $Y=y$ in $(M,\vec{u})$ according to a definition that uses minimal necessity, then $\vec{X}$ is a singleton.
\eopro

\prf 
Since we know that {\bf Def 7} is unsatisfiable and we have Theorem \ref{thm:ident}, we only need to consider {\bf Def 3}, {\bf Def 8}, and {\bf Def 10}. The following applies to both weak and direct sufficiency (i.e., {\bf Def 3} and {\bf Def 10}.)

Assume $(\vec{X_1}=\vec{x_1},\vec{X_2}=\vec{x_2},\vec{W}=\vec{w}^*)$ is sufficient for $Y=y$, and $\vec{W}=\vec{w}^*$ is not sufficient for $Y=y$. If either $(\vec{X_2}=\vec{x_2},\vec{W}=\vec{w}^*)$ or $(\vec{X_1}=\vec{x_1},\vec{W}=\vec{w}^*)$ is also sufficient for $Y=y$, then $(\vec{X_1}=\vec{x_1},\vec{X_2}=\vec{x_2})$ is not minimal. 

So let us assume that neither $(\vec{X_2}=\vec{x_2},\vec{W}=\vec{w}^*)$ nor $(\vec{X_1}=\vec{x_1},\vec{W}=\vec{w}^*)$ is sufficient for $Y=y$. This means we can move $\vec{X_2}$ to the witness to show that $\vec{X_1}=\vec{x_1}$ satisfies AC2 by itself, and likewise for $\vec{X_2}$ and $\vec{X_1}$ reversed. From this the result follows.

Now we prove that it also holds for strong sufficiency, i.e., for {\bf Def 8}. Assume $(\vec{X_1}=\vec{x_1},\vec{X_2}=\vec{x_2},\vec{W}=\vec{w}^*)$ is sufficient for $Y=y$ along $\vec{N}$, and $\vec{W}=\vec{w}^*$ is not sufficient for $Y=y$ along any network $\vec{S} \subseteq \vec{N}$. If either $(\vec{X_2}=\vec{x_2},\vec{W}=\vec{w}^*)$ or $(\vec{X_1}=\vec{x_1},\vec{W}=\vec{w}^*)$ is also sufficient for $Y=y$ along $\vec{N}$, then $(\vec{X_1}=\vec{x_1},\vec{X_2}=\vec{x_2})$ is not minimal. 

So let us assume that neither $(\vec{X_2}=\vec{x_2},\vec{W}=\vec{w}^*)$ nor $(\vec{X_1}=\vec{x_1},\vec{W}=\vec{w}^*)$ is sufficient for $Y=y$ along $\vec{N}$. If the same is true for all subnetworks $\vec{S} \subseteq \vec{N}$, then as before, we can move either one of $\vec{X_1}$ and $\vec{X_2}$ to the witness to show that the other satisfies AC2 by itself. 

So let us assume that there is some subnetwork $\vec{S}' \subseteq \vec{N}$ such that $(\vec{X_1}=\vec{x_1},\vec{W}=\vec{w}^*)$ is sufficient for $Y=y$ along $\vec{S}'$. (Obviously the same reasoning applies to $\vec{X_2}$.) Since all subnetworks $\vec{S}''$ of $\vec{S}'$ are also subnetworks of $\vec{N}$, it follows from the above that $(\vec{X_1}=\vec{x_1})$ satisfies AC2 by itself when taking $\vec{W}$ as witness and $\vec{S}'$ as network. From this the result follows.
\eprf

\othm{thm:relations} The only implications -- involving either causes or parts of causes -- between the remaining five definitions ({\bf Def 2}, {\bf Def 3}, {\bf Def 4}, {\bf Def 8}, and {\bf Def 10}) and the three HP definitions are the following ones (and their immediate consequences, of course):
\begin{itemize}
	\item If part of {\bf Modified HP} then {\bf Updated HP};
	\item If part of {\bf Updated HP} then {\bf Original HP}; 
	\item If {\bf Def 3} then {\bf Def 2};
	\item If part of {\bf Def 2} then {\bf Def 8};
	\item If {\bf Def 3} then {\bf Original HP};
	\item If {\bf Def 10} then {\bf Def 4}. 
\end{itemize}
\eothm

\prf The first two implications are proven in \citep{halpernbook}.

First we prove the third implication. Assume $\vec{X}=\vec{x}$ causes $Y=y$ with witness $\vec{W}$ according to {\bf Def 3}. It follows from Proposition \ref{pro:single} that $\vec{X}$ is a single conjunct $X$. Note that this immediately implies minimality of $\vec{X}$. 

In other words, $(X=x,\vec{W}=\vec{w}^*)$ is directly sufficient for $Y=y$, and there exists some $x'$ such that $(X=x',\vec{W}=\vec{w}^*)$ is not directly sufficient for $Y=y$. From the former it follows that $(X=x,\vec{W}=\vec{w}^*)$ is strongly sufficient for $Y=y$ along $\emptyset$. From the latter it follows that $(X=x',\vec{W}=\vec{w}^*)$ is not strongly sufficient for $Y=y$ along $\emptyset$, from which the result follows.

Second we prove the fourth implication. Assume $(X=x,\vec{X_2}=\vec{x_2},\vec{W}=\vec{w}^*)$ is sufficient for $Y=y$ along $\vec{N}$, and $(X=x',\vec{X_2}=\vec{x_2}',\vec{W}=\vec{w}^*)$ is not sufficient for $Y=y$ along any network $\vec{S} \subseteq \vec{N}$, for some $\vec{N}$, $x'$ and $\vec{x_2}'$. We show that $X=x$ causes $Y=y$ according to {\bf Def 8}. 

Taking $(\vec{X_2}=\vec{x_2},\vec{W}=\vec{w}^*)$ as our witness and using $\vec{N}$, AC2(b) remains unchanged. If $(\vec{X_2}=\vec{x_2},\vec{W}=\vec{w}^*)$ is not sufficient for $Y=y$ along any network $\vec{S} \subseteq \vec{N}$, then the result follows. We proceed by a reductio. 

Let us assume that $(\vec{X_2}=\vec{x_2},\vec{W}=\vec{w}^*)$ is sufficient for $Y=y$ along some $\vec{S} \subseteq \vec{N}$. If $(\vec{X_2}=\vec{x_2}',\vec{W}=\vec{w}^*)$ is not sufficient for $Y=y$ along any $\vec{S}'' \subseteq \vec{S}$, we have a violation of minimality (since $X$ is redundant). Therefore we know that $(\vec{X_2}=\vec{x_2}',\vec{W}=\vec{w}^*)$ is sufficient for $Y=y$ along some network $\vec{S}'' \subseteq \vec{S}$. 

This means that there exist values $\vec{s}'' \in \R(\vec{S}'')$ so that for all settings $\vec{c} \in \R(\V \minus (\vec{S}'' \cup \vec{X_2} \cup \{X,Y\})$, and for all $x'' \in \R(X)$, it holds that $(M,\vec{u})\sat[\vec{X_2} \gets \vec{x_2}', \vec{W} \gets \vec{w}^*,\vec{C} \gets \vec{c}, X \gets x'']\vec{S}=\vec{s}''$ and $(M,\vec{u})\sat[\vec{X_2} \gets \vec{x_2}', \vec{W} \gets \vec{w}^*,\vec{C} \gets \vec{c}, X \gets x'',\vec{S} \gets \vec{s}'']Y=y$. In particular, this holds if we choose $X=x'$. But that means that $(X=x',\vec{X_2}=\vec{x_2}',\vec{W}=\vec{w}^*)$ is also sufficient for $Y=y$ along $\vec{S}''$, which contradicts our starting assumption. 

Third we prove the fifth implication. As with the third implication, assume that $(X=x,\vec{W}=\vec{w}^*)$ is directly sufficient for $Y=y$, and there exists some $x'$ such that $(X=x',\vec{W}=\vec{w}^*)$ is not directly sufficient for $Y=y$. From the latter it follows that there exists a setting $\vec{d}$ of $\V \minus (\vec{X} \cup \vec{W} \cup \{Y\})$ such that $(M,\vec{u}) \sat [X \gets x',\vec{W} \gets \vec{w}^*, \vec{D} \gets \vec{d}] Y \neq y$. This means that if we take $(\vec{W}=\vec{w}^*,\vec{D}=\vec{d})$ as witness, AC2(a) is satisfied for {\bf Original HP}. Since $(X=x,\vec{W}=\vec{w}^*)$ is directly sufficient for $Y=y$, we know that $(M,\vec{u}) \sat [X \gets x,\vec{W} \gets \vec{w}^*, \vec{D} \gets \vec{d}] Y = y$. Also, we have that $\vec{Z}=\vec{X}$, and thus the former means that also AC2(b) is satisfied for {\bf Original HP}.

Fourth we prove the last implication.  Assume $X=x$ causes $Y=y$ with witness $\vec{W}$ according to {\bf Def 10}. (We know because of Proposition \ref{pro:single} that $\vec{X}$ is a singleton.) In other words, $(X=x,\vec{W}=\vec{w}^*)$ is weakly sufficient for $Y=y$, and $\vec{W}=\vec{w}^*$ is not weakly sufficient for $Y=y$. Remains to be shown that there exist a value $x'$ so that $(X=x',\vec{W}=\vec{w}^*)$ is not weakly sufficient for $Y=y$. 

Say $\vec{u}'$ is a context such that $(M,\vec{u}') \sat [\vec{W} \gets \vec{w}^*] Y \neq y$, and say $x'$ is the unique value such that $(M,\vec{u}') \sat [\vec{W} \gets \vec{w}^*] X = x'$. Then also $(M,\vec{u}') \sat [X \gets x',\vec{W} \gets \vec{w}^*] Y \neq y$, which is what remained to be shown.

Fifth, we show that none of the remaining implications hold. (Again, we do not consider the relations amongst the HP definitions explicitly and refer the reader to the examples in Section \ref{sec:237}. We also do not explicitly consider the remaining implications for parts of causes, but the reader can verify that the following examples suffice to falsify all those implications as well. For the left-hand side of all implications this follows immediately from the fact that the causes in all the following examples are singletons. For the right-hand side of implications, Propositions \ref{pro:single}, \ref{pro:single2}, and \ref{pro:parent} come in handy.)

Example \ref{ex2} shows that {\bf Def 4} does not imply {\bf Def 10}.

Example \ref{ex1} shows that none of the other definitons imply either {\bf Def 4} or {\bf Def 10}. So there are no remaining implications with either {\bf Def 4} or {\bf Def 10} on the right-hand side.

Example \ref{ex4} shows that {\bf Def 3} is not implied by any definition. 

Example \ref{ex3} shows that none of the HP definitions imply {\bf Def 2} or {\bf Def 8}. Note that {\bf Def 4} and {\bf Def 10} also consider $X=1$ a cause of $Y=1$ in that example (since $X=1$ is weakly sufficient for $Y=1$, whereas $X=0$ or the emptyset is not). Further, Example \ref{switch} shows that {\bf Def 8} does not imply {\bf Def 2}. Therefore there are no remaining implications with {\bf Def 2} or {\bf Def 8} on the right-hand side.

That leaves us to consider implications with one of the HP definitions on the right-hand side. Given the first two implications of Theorem \ref{thm:relations}, it suffices to show that none of {\bf Def 4}, {\bf Def 2}, {\bf Def 8}, or {\bf Def 10}, imply {\bf Original HP}, and that {\bf Def 3} does not imply {\bf Updated HP}.

I refer the reader to Example \ref{counter} in Section \ref{sec:237} for an example where {\bf Def 2} -- and thus also {\bf Def 8} -- hold and {\bf Original HP} does not.

The following example shows that neither {\bf Def 4} nor {\bf Def 10} implies {\bf Original HP}.
\begin{example}\label{ex5} Equations: $Y=Z_1 \lor Z_2 \lor A$, $Z_1=X \land A$, $Z_2=X \land \lnot A$. Context: $A=1$ and $X=1$. Then $X=1$ is a cause of $Y=1$ according to: 
	\begin{itemize}
		\item {\bf Def 10}: $X=1$ is weakly sufficient for $Y=1$ and $\emptyset$ is not.
		\item {\bf Def 4}: follows from the previous one. 
	\end{itemize}
	
Yet $X=1$ is not a cause of $Y=1$ according to {\bf Original HP}. To see why, note that we need to include $A=0$ into the witness in order to get AC2(a), and we must exclude $Z_1$. Also, we clearly cannot add $Z_2=1$. Therefore the witness has to be $A=0$. The actual value of $Z_2$ is $0$. Since we have $(M,\vec{u}) \sat [X \gets 1, A \gets 0, Z_2 \gets 0]Y=0$, AC2(b) is not satisfied.
\end{example}

Lastly, an example to show that {\bf Def 3} does not imply {\bf Updated HP}.
\begin{example}\label{ex6} Equations:  $Y=(X \land D) \lor A$, $D=A$. Context: $A=1$ and $X=1$. Then $X=1$ is a cause of $Y=1$ according to {\bf Def 3}: $(X=1,D=1)$ is directly sufficient for $Y=1$, and $(X=0,D=1)$ is not. But $X=1$ is not a cause of $Y=1$ according to {\bf Updated HP}. To see why, note that we need to include $A=0$ into the witness in order to get AC2(a). But $(M,\vec{u}) \sat [X \gets 1, A \gets 0]Y=0$, thus falsifying AC2(b) for {\bf Updated HP}.
\end{example}

\eprf

\section*{Excluding {\bf Def 3} and {\bf Def 10}}

\opro{pro:single2} If $\vec{X}=\vec{x}$ causes $Y=y$ in $(M,\vec{u})$ according to {\bf Def 3}, then $\vec{X}$ is a singleton, and $X$ is a parent of $Y$. 
\eopro

\prf That $\vec{X}$ is always a singleton is a direct consequence of the combination of Proposition \ref{pro:single} and Theorem \ref{thm:ident}.

Recall that $X$ is a parent of $Y$ iff there exists a context $\vec{u}''$, a setting $\vec{z} \in \R(\V \minus \{X,Y\})$, and values $x,x''$ of $X$ so that $F_Y(\vec{u}'',\vec{z},x) \neq F_Y(\vec{u}'',\vec{z},x'')$. This means precisely that for some $y \in \R(Y)$, $(M,\vec{u}'') \sat [\vec{Z} \gets \vec{z}, X \gets x]Y=y$ and $(M,\vec{u}'') \sat [\vec{Z} \gets \vec{z}, X \gets x'']Y \neq y$. If $X=x$ causes $Y=y$ according to {\bf Def 3}, the existence of values such that the previous holds follows immediately.
\eprf

\opro{pro:parent} If $X$ is only a parent of $Y$, then {\bf Def 3}, {\bf Def 2}, and {\bf Def 8} are all equivalent for causes $X=x$.
\eopro

\prf Given Theorem \ref{thm:relations}, we only need to prove the implication from {\bf Def 8} to {\bf Def 3}.

Assume $X$ is only a parent of $Y$, and $X=x$ causes $Y=y$ according to {\bf Def 8}. Thus, there is a witness $\vec{W}$ and some network $\vec{N}$ such that $(X=x,\vec{W}=\vec{w}^*)$ is strongly sufficient for $Y=y$ along $\vec{N}$, and $(\vec{W}=\vec{w}^*)$ is not strongly sufficient for $Y=y$ along any subnetwork of $\vec{N}$. 

First consider the case where $\vec{N}=\emptyset$. This means that $(X=x,\vec{W}=\vec{w}^*)$ is directly sufficient for $Y=y$, and $(\vec{W}=\vec{w}^*)$ is not directly sufficient for $Y=y$. That means precisely that $X=x$ causes $Y=y$ according to {\bf Def 12}. The result now follows from Theorem \ref{thm:ident}. 

Second consider the case where there exists some $N \in \vec{N}$. If $N$ is not an ancestor of $Y$, it can be removed from $\vec{N}$ without consequence. If $N$ is an ancestor of $Y$, then it cannot be a descendant of $X$. But in that case it does not depend on $X$, and thus we can remove it from $\vec{N}$ and add it to the witness $\vec{W}$ without consequence. Therefore there always exists a choice of witness so that $\vec{N}=\emptyset$, and thus the result follows.
\eprf

\opro{pro:dep} Out of all definitions we have considered, {\bf Def 10} and {\bf Def 3} are the only ones which do not satisfy {\bf Dependence}.
\eopro 

\prf
For the HP definitions this is proven in \citep[p. 26]{halpernbook}.

Example \ref{ex4} shows the result for {\bf Def 3}.

Example \ref{ex2} shows the result for {\bf Def 10}.

Therefore it remains to be shown that {\bf Dependence} implies {\bf Def 2}, {\bf Def 4}, and {\bf Def 8}. This is a direct consequence of the fact that {\bf Dependence} implies {\bf Modified HP}, combined with Proposition \ref{pro:mod}. 

\eprf

\section*{{\bf Def 2}, {\bf Def 4}, and {\bf Def 8}, vs the HP definitions}

\opro{pro:mod} If {\bf Modified HP} with $\vec{X}$ a singleton, then {\bf Def 2}, {\bf Def 4}, and {\bf Def 8}.
\eopro

\prf
Recall the root variables $\vec{R}$ from Observation \ref{obs:obs1}. Note that for any setting $\vec{r} \in \R(\vec{R})$, for any set $\vec{Y} \subseteq (\V \minus \vec{R})$, there exists some $\vec{y}$ so that $\vec{R}=\vec{r}$ is both weakly, actually weakly, and strongly, sufficient for $\vec{Y}=\vec{y}$.  

Assume $X=x$ causes $Y=y$ according to {\bf Modified HP} with witness $\vec{W}$. This means there exists a $x'$ so that $(M,\vec{u})\sat [X \gets x', \vec{W} \gets \vec{w}^*] Y \neq y$. Let $\vec{S}=\vec{R} \minus (\vec{W} \cup \{X\})$. 

First we focus on {\bf Def 4}. Note that $(X=x,\vec{S}=\vec{s}^*,\vec{W}=\vec{w}^*)$ is weakly sufficient for $Y=y$. Furthermore, changing $X$ from $x$ to $x'$ obviously has no effect on any of the values in $\vec{R}$. Therefore $(M,\vec{u})\sat [X \gets x', \vec{W} \gets \vec{w}^*] \vec{S}=\vec{s}^*$, and thus we get that $(M,\vec{u})\sat [X \gets x', \vec{W} \gets \vec{w}^*,\vec{S} \gets \vec{s}^*] Y \neq y$. (Also, we may assume that $\vec{W} \cap \vec{R}=\emptyset$.) From this it follows that $(X=x',\vec{S}=\vec{s}^*,\vec{W}=\vec{w}^*)$ is not weakly sufficient for $Y=y$. So taking $(\vec{S}=\vec{s}^*,\vec{W}=\vec{w}^*)$ as witness gives the desired result.

Second we focus on {\bf Def 2} (from which {\bf Def 8} follows due to Theorem \ref{thm:relations}). Combining the previous statement about $(X=x',\vec{S}=\vec{s}^*,\vec{W}=\vec{w}^*)$  with Proposition \ref{pro:relat} it follows immediately that there does not exist any network $\vec{N}$ so that $(X=x',\vec{S}=\vec{s}^*,\vec{W}=\vec{w}^*)$ is strongly sufficient for $Y=y$ along $\vec{N}$. 

Clearly there exists some $\vec{N}$ so that $\vec{R}=\vec{r}^*$ is strongly sufficient for $Y=y$ along $\vec{N}$. (We can start by picking parents $\vec{A}$ of $Y=y$ such that $\vec{A}=\vec{a}^*$ is directly sufficient for $Y=y$. Then we can take parents of all elements in $\vec{A}$, to get a set $\vec{B}$ so that $\vec{B}=\vec{b}^*$ is directly sufficient for $\vec{A}=\vec{a}^*$, etc.) But then also $(X=x,\vec{S}=\vec{s}^*,\vec{W}=\vec{w}^*)$ is strongly sufficient for $Y=y$ along $\vec{N}$, from which the result follows.
\eprf

}

\section*{Acknowledgements}
Many thanks to Joe Halpern and Naftali Weinberger for helpful comments on earlier versions of this paper. This research was made possible by funding from the Alexander von Humboldt Foundation.

\bibliographystyle{spbasic} 
\bibliography{allpapers}

\end{document}